\documentclass[12pt]{iopart}
\usepackage[utf8]{inputenc}

\usepackage{cite}
\usepackage{adjustbox}
\usepackage{multirow}
\usepackage{graphicx}
\usepackage{float,lscape}
\usepackage{rotating}
\usepackage{caption}
\usepackage{subcaption}
\usepackage{multicol}
\usepackage{threeparttablex}

\usepackage{tikz}
\usetikzlibrary{shapes,decorations}

\newcommand{\shapelabel}[3]{
    \begin{tikzpicture}[baseline={([yshift=-0.8ex]current bounding box.center)}]
    \protect\node[#1,fill=#2,draw=black,scale=.8]{\color{white}\textbf{#3}};
    \end{tikzpicture}
}

\tikzset{
    shapelabel/.style n args={2}{
        #1,
        draw=black,
        fill=#2,
        baseline={([yshift=-0.8ex]current bounding box.center)},
        text=white
    }
}

\def\BibTeX{{\rm B\kern-.05em{\sc i\kern-.025em b}\kern-.08em
    T\kern-.1667em\lower.7ex\hbox{E}\kern-.125emX}}
\begin{document}

\title[Static Hand Gesture Recognition for American Sign Language using Neuromorphic]{Static Hand Gesture Recognition for American Sign Language using Neuromorphic Hardware}

\author{MohammadReza Mohammadi$^1$, Peyton Chandarana$^1$, James Seekings, Sara Hendrix, Ramtin Zand}
\address{University of South Carolina, Columbia, SC 29201}
\ead{ramtin@cse.sc.edu}
\vspace{10pt}

\begin{indented}
\item[]\today
\vspace{10pt}
\item[]$^1$Authors contributed equally.
\end{indented}
\begin{abstract}
In this paper, we develop four spiking neural network (SNN) models for two static American Sign Language (ASL) hand gesture classification tasks, i.e., the ASL Alphabet and ASL Digits. The SNN models are deployed on Intel's neuromorphic platform, Loihi, and then compared against equivalent deep neural network (DNN) models deployed on an edge computing device, the Intel Neural Compute Stick 2 (NCS2). We perform a comprehensive comparison between the two systems in terms of accuracy, latency, power consumption, and energy. The best DNN model achieves an accuracy of 99.93\% on the ASL Alphabet dataset, whereas the best performing SNN model has an accuracy of 99.30\%. For the ASL-Digits dataset, the best DNN model achieves an accuracy of 99.76\% accuracy while the SNN achieves 99.03\%. Moreover, our obtained experimental results show that the Loihi neuromorphic hardware implementations achieve up to 20.64$\times$ and 4.10$\times$ reduction in power consumption and energy, respectively, when compared to NCS2.

\end{abstract}
%
%
%
%
%

\section{Introduction}

Sign language is a visual language that enables people who are deaf or hard of hearing to communicate with others in their communities. To convey emotion, grammar, and sentence structure similar to spoken language, sign language employs visual and manual elements such as hand gestures, facial expressions, and body movements. Hand gestures are regarded as the fundamental component of a sign language vocabulary. In addition to these gestures, facial expressions and body movements are used to accentuate the emotions of words and phrases\cite{cheok2019review}. Hand gestures in a sign language can be classified as static or dynamic depending on whether or not hand motion is incorporated into the sign interpretations \cite{tolentino2019static , liao2019dynamic}. Static hand gestures, also known as static hand postures, consist of the shape and orientation of the hand and fingers and are commonly used for fingerspelling alphabet letters and digits. Dynamic hand gestures, on the other hand, are a set of hand gestures accompanied by motion for word interpretation and translation. 
\par
Automatic hand gesture recognition is applicable to a variety of applications and has been a broad and significant research area over the last decade \cite{wadhawan2021sign}. The most widely researched approaches to automatic hand gesture recognition employ motion sensors and vision sensors. 
\par
Motion sensor techniques collect and track hand gestures, finger orientation, and hand velocity using a range of sensors such as 
sensor gloves \cite{shukor2015new}, motion controllers \cite{kumar2017real}, accelerometers \cite{zhang2011framework}, magnetometers, and gyroscopes \cite{hernandez2004new,bui2007recognizing}.
\textcolor{black}{Many works in the past have adopted multiple motion sensors simultaneously to increase the classification accuracy of the performed signs \cite{elbadawy2015proposed, 7025313,kumar2017coupled}.}
Employing multiple motion sensors in a recognition system allows for a more precise recognition in particular sign language vocabularies but consequentially results in higher implementation cost and complexity.
\textcolor{black}{While motion sensor-based approaches have the benefit of being able to accurately capture motion data in real-time,
these sensors are typically expensive and inconvenient for signers since installing or wearing these instruments can restrict motion and thus hinder performance.}
\par
On the other hand, vision-based approaches utilize at least one camera and image processing algorithms including segmentation, shape detection, motion tracking, color detection, and contour modeling to capture and recognize actual hand gestures without the need for cumbersome physical devices \cite{garcia2016real, rautaray2015vision,pisharady2015recent,d2016recent}. Since these approaches do not require wearing the devices to track motion, they are more user-friendly and comfortable without impeding the wearer's motions. Similar to many other object detection and recognition applications, deep learning methods, particularly convolutional neural networks (CNNs), have recently become the primary choice for sign language applications \cite{rastgoo2018multi,adithya2020deep,barbhuiya2021cnn,rahman2019new}. 
However, the major downside of deep learning-based techniques is that they often consume a significant amount of power \cite{verhelst2017embedded}. Since hand gesture recognition models are typically intended to be deployed on portable devices like smartphones, tablets, and wearable devices like smart glasses, power and energy are major constraints. Low-power machine learning is an important field of study that can enable such applications at the edge devices. In particular, neuromorphic algorithms and hardware, which are inspired by the low-power processing of the human brain, have been attracting more attention in recent years \cite{schuman2022opportunities, loihi_2021}. 
\par
In 2018, Intel introduced a neuromorphic processor called Loihi capable of running spiking neural networks (SNNs) with orders of magnitude lower power consumption while achieving accuracies close to state-of-the-art CPU, GPU, and TPU implementations \cite{loihi_2018}. Along with Loihi, Intel provided a python library, NxSDK, for SNN design and deployment. NxSDK included many features for designing SNNs from scratch including setting learning rules for synaptic plasticity learning methods. This design methodology, however, does introduce its own complexities in design, training, and hyper-parameter tuning to get accurate and meaningful results.
\par
While there are inherently many differences between how artificial neural networks (ANNs) and SNNs are trained and operate, conversion processes have been widely researched to convert ANNs to SNNs to reduce the complexity of the end-to-end design of SNNs. In 2017, Rueckauer et. al.\cite{snntoolbox} developed the SNN Conversion Toolbox which aimed to convert trained ANN models to SNNs. While the concept of converting ANNs to SNNs was not novel at the time with many works introducing different conversion methods \cite{perez_conversion,cao_conversion}, the SNN Conversion Toolbox intended to streamline the process for conversion by both creating a more automatic conversion pipeline and also implementing several features missing in the previous conversion works. 
\textcolor{black}{Unlike previous conversion methods, the SNN Conversion Toolbox allows an ANN to be trained in a deep learning library like TensorFlow \cite{tensorflow} or PyTorch \cite{pytorch} with layers commonly found in CNN architectures.}
The toolbox implements spiking layers like average pooling and convolutional layers to provide a means for the models to be parsed and converted to SNNs without as much hyperparameter tuning and as many design considerations.
In 2021, Rueckauer et. al. \cite{nxtf} introduced NxTF which extended the SNN Conversion Toolbox to enable the deployment of converted SNNs to Intel's Loihi neuromorphic chip. NxTF was designed specifically to map layers found in the ANN models to layers that Loihi could understand through a custom NxSDK backend \cite{loihi_2021}.
\par
In this paper, we focus on classifying static images of the ASL Alphabet and ASL Digits. The following are our contributions:
\begin{itemize}
\item We design four ANN models and train each model with the ASL Alphabet and ASL Digit static image datasets for a total of eight models. We then convert them to SNNs using the SNN Conversion Toolbox.
\item We analyze the trade-off between accuracy and latency and compare the differences when favoring accuracy or latency.
\item We investigate methods for accurately measuring power and energy consumption.
\item We compare and analyze the hardware performance of the ANN models on the Intel Neural Compute Stick 2 to the SNN models on Intel Loihi in terms of accuracy, latency, power consumption, and energy.
\end{itemize}
\par
\textcolor{black}{The subsequent sections of this paper are organized as follows.} The datasets for this study are presented in Section 2. Section 3 describes the ANN model design. Section 4 provides a summarized background on the SNN Conversion Toolbox's conversion methodology. Section 5 introduces the experimental methodology performed in this work including the processes for hardware deployment and measuring latency, power, and energy. The results of the experiments are discussed and evaluated in Section 6. Section 7 concludes the paper and provides some discussion for future study.
\section{ASL Datasets}


\begin{figure}
    \begin{subfigure}[b]{.505\linewidth}
        \includegraphics[width=\textwidth]{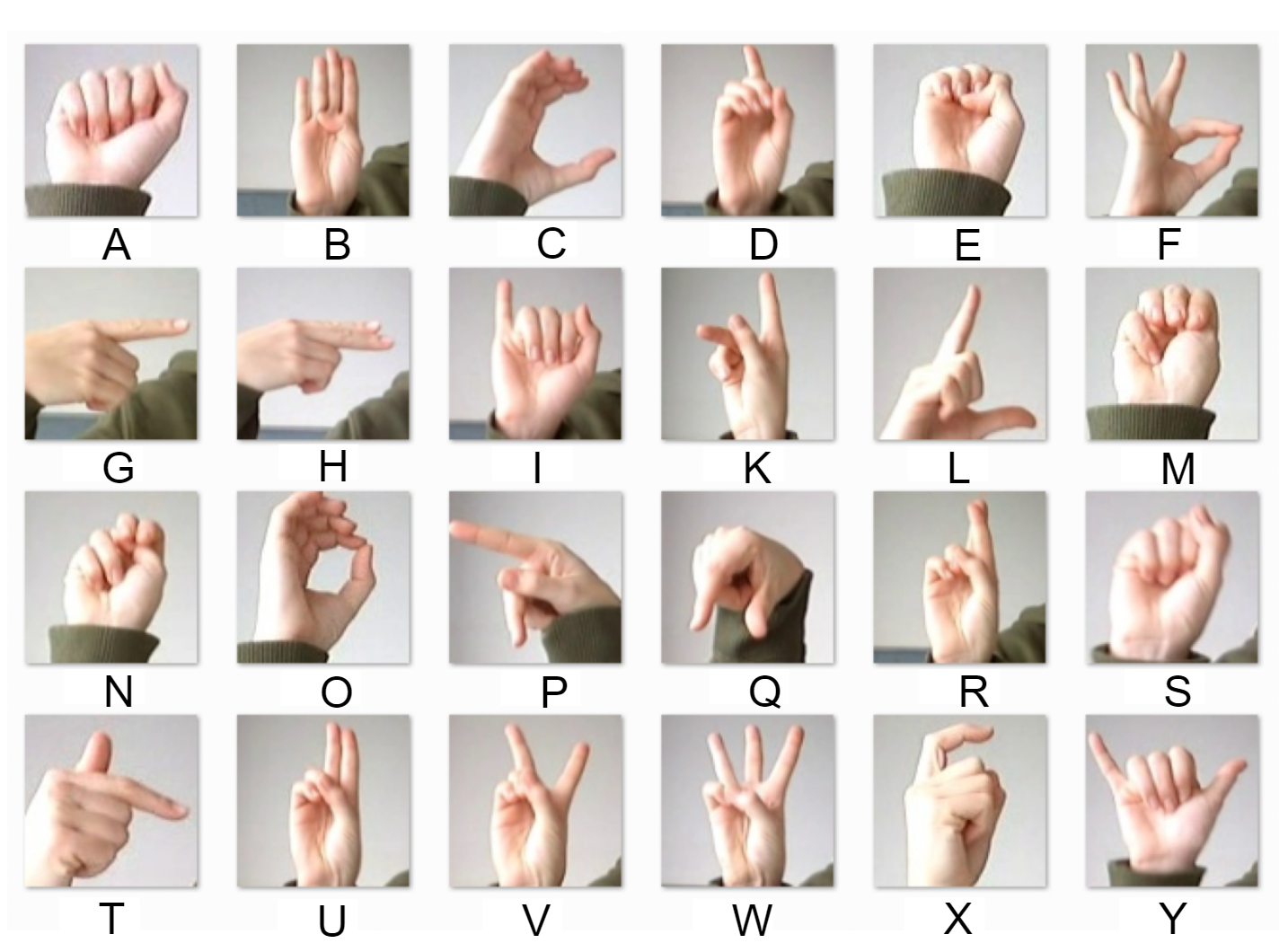}
        \caption{Samples images of ASL Alphabet \cite{signlangmnistkaggle}}
        \label{fig:y equals x}
    \end{subfigure}
    \begin{subfigure}[b]{.545\linewidth}
        \includegraphics[width=\textwidth]{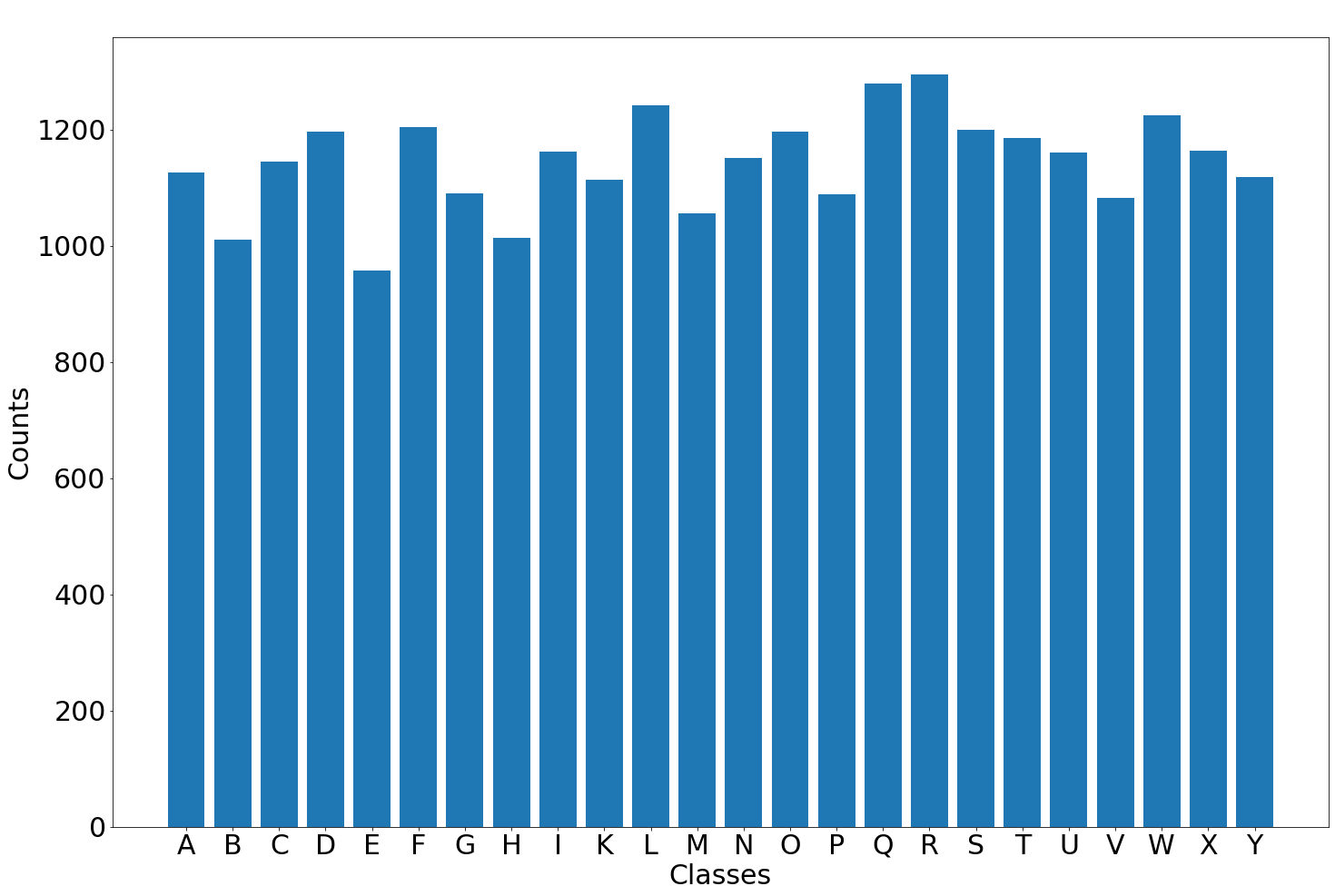}
        \caption{ASL Alphabet Distribution}
        \label{fig:three sin x}
    \end{subfigure}
    \begin{subfigure}[b]{\linewidth}
        \centering
        \includegraphics[width=.6\linewidth]{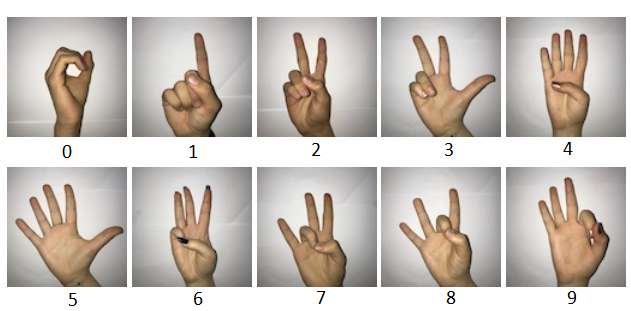}
        \caption{Sample images of ASL Digits \cite{mavi2020new}}
        \label{fig:five over x}
    \end{subfigure}
    \caption{Samples of the ASL images used in classification. The ASL Digit dataset has equal class distribution while the ASL Alphabet does not. \textcolor{black}{While the ASL Alphabet distribution is uniform, it is not skewed to favor one or multiple classes.}}
    \label{fig:sample}
\end{figure}

In this paper, two different static image datasets are studied. The American Sign Language (ASL) Alphabet \cite{signlangmnistkaggle} is the first dataset and replaces the common handwritten digit dataset, MNIST \cite{lecun2010mnist}, commonly used as an ANN model's proof-of-concept dataset in classification. The ASL Alphabet dataset 
includes static images of multiple people repeating ASL finger-spelling against various backdrops. With the exception of J and Z, which require motion, the ASL Alphabet dataset of hand gestures is a multi-class problem with 24 classes of letters. Figure \ref{fig:sample}(a) depicts examples of each class. The ASL Alphabet dataset, created by \cite{signlangmnistkaggle}, was constructed by considerably expanding a small set of 1704 color images that were not cropped around the hand region of interest. The designers of the dataset, \cite{signlangmnistkaggle}, generate additional data using an image processing pipeline which is comprised of cropping the images around the hands, converting the images to gray-scale, resizing the images, and then making more than 50 variations to increase the dataset size. Furthermore, the data augmentation includes the use of Mitchell, Hermite, and Catrom filters, coupled with 5\% random noise, \textpm 15\% brightness/contrast, and 3 degrees of image rotation. After these data augmentation steps, the dataset includes 34,627 gray-scale images of size 28$\times$28 pixels which are split into the separate train, validation, and test datasets. 27,455 samples, $\sim$80\%, of the ASL alphabet are used for training and validation, and 7,172 samples, $\sim$20\%, are used for testing. Figure \ref{fig:sample}(b) shows the distribution of training samples across 24 classes.

\par The other dataset we employ in this study is the ASL Digits dataset \cite{mavi2020new}. It is comprised of 2062 RGB images with 100$\times$100 pixels that are divided into 10 classes, digits 0-9. We resized the resolution of these images to 28$\times$28 and converted the RGB images to gray-scale. 20\% of this dataset is used as the test dataset (413 images) and the remaining images are used for validation, 330 images, and training. Figure \ref{fig:sample} (c) shows a sample image for each class.

\section{Model Design}
We use three CNN models and a multi-layer perceptron (MLP) neural network for the static ASL image classification task. The CNN models are inspired by three standard and well-known models, i.e., LeNet, AlexNet, and VGGNet. Compared to the original models, we slightly modified our models to create smaller versions that still achieve high accuracy values. Moreover, we constrained the models such that they could be readily converted to SNN models. The specifics for each implementation are provided in the following subsections.


\subsection{MLP}
Figure \ref{fig:MLP} shows the structure of the MLP that we used in this work. \textcolor{black}{The proposed MLP contains two hidden layers with 512 and 256 neurons each, as well as 24 or 10 output neurons for the ASL Alphabet or ASL Digits dataset, respectively.} The first and second layers are followed by a dropout layer each with a probability of 0.2. We use  the ReLU activation function for the hidden layers and the softmax activation function for the output layer.

\subsection{LeNet}

\begin{figure}
     \begin{subfigure}[b]{0.36\textwidth}
         \centering
         \includegraphics[width=\textwidth]{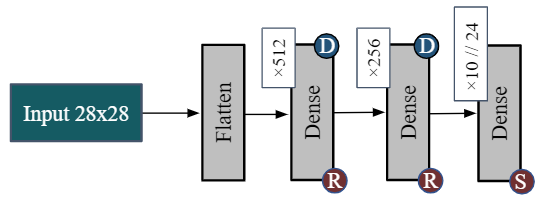}
         \caption{$MLP$}
         \label{fig:MLP}
     \end{subfigure}
     \begin{subfigure}[b]{0.63\textwidth}
         \centering
         \includegraphics[width=\textwidth]{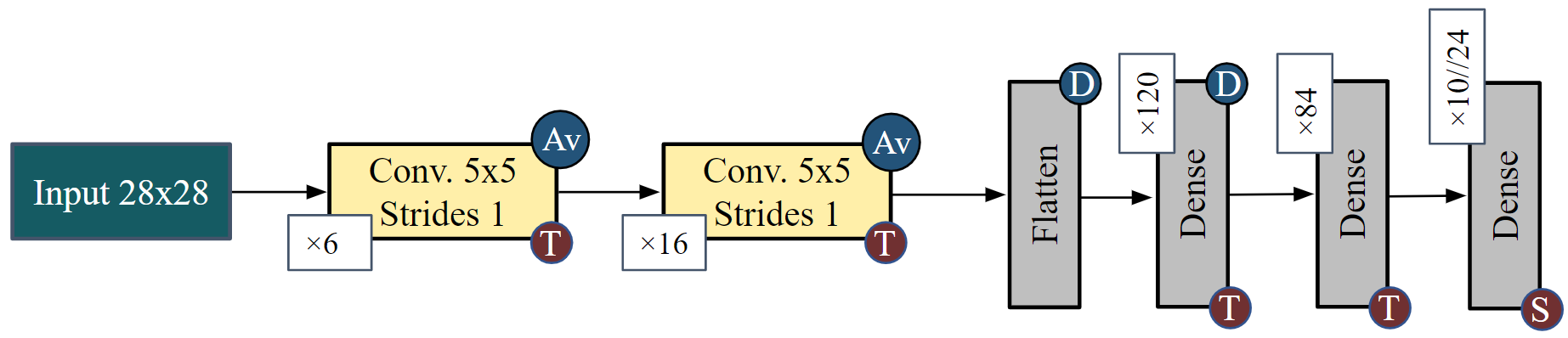}
         \caption{\textit{LeNet-5}}
         \label{fig:Lenet}
     \end{subfigure}
     \hfill
    \begin{center}
     \begin{subfigure}[b]{0.85\textwidth}
         \centering
         \includegraphics[width=\textwidth]{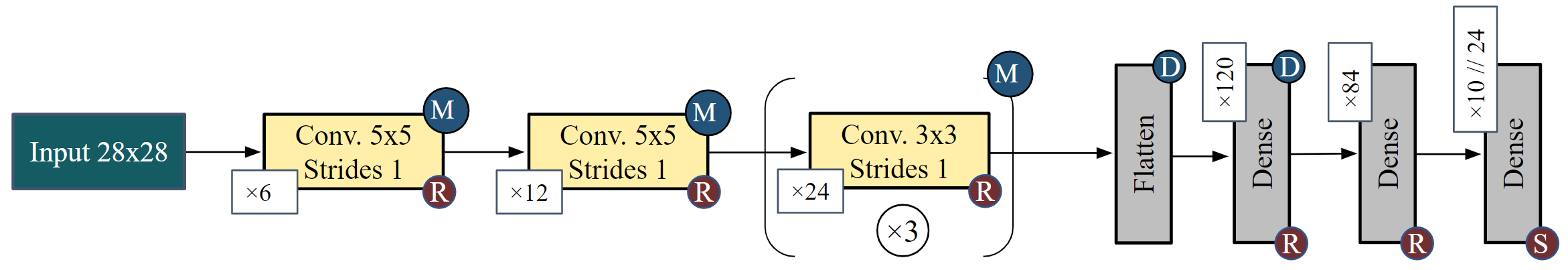}
         \caption{$AlexNet$}
         \label{fig:Alex}
     \end{subfigure}
     \end{center}
     \hfill
    \begin{center}
     \begin{subfigure}[b]{1\textwidth}
         \centering
         \includegraphics[width=\textwidth]{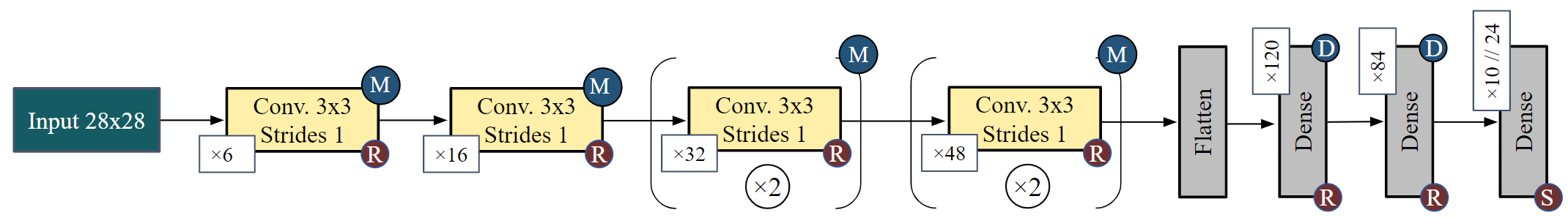}
         \caption{\textit{VGG-9}}
         \label{fig:VGG}
     \end{subfigure}
     \end{center}
        \caption{ANN architectures for the ASL Alphabet (24 output classes) and ASL Digits (10 output classes) classification.
        \shapelabel{circle}{black!60!red}{R} ReLU, \shapelabel{circle}{black!60!red}{T} Tanh, \shapelabel{circle}{blue!60!green}{M} MaxPooling2D, \shapelabel{circle}{blue!60!green}{Av} AveragePooling2D, \shapelabel{circle}{blue!60!green}{D} Dropout, \shapelabel{circle}{black!60!red}{S} Softmax.}
        \label{fig:annarchs}
\end{figure}

The architecture of the Lenet \cite{lecun1998gradient} model that we employed here is shown in figure \ref{fig:Lenet}. It is comprised of two convolutional layers and three fully connected layers. The first and second convolution layers consist of 6 and 16 kernels of size 5$\times$5, respectively. After the first and second convolution layers, there is a non-overlapping average pooling layer with a 2$\times$2 filter size and strides of 2. The three fully-connected layers each include 120, 84, and 24 or 10 neurons depending on which dataset is used. In this study, we modified the original LeNet model and applied two dropout layers with a probability of 0.25 to prevent overfitting after the first and the second fully-connected layers. 


\subsection{AlexNet}
The AlexNet model \cite{krizhevsky2012imagenet} consists of five convolutional layers each followed by max pooling and 3 fully connected (FC) layers, all of which utilize the ReLU activation function except the output layer, which uses the softmax activation function. The original AlexNet model is quite large, with over 62 million parameters. Here, we changed the structure of the original network as follows. Instead of 96 11$\times$11 kernels with a stride of 4, just 6 5$\times$5 kernels with a stride of 1 are used for the first convolution layer. Instead of 256 kernels, we employ 12 5$\times$5 kernels for the second convolution layer. We utilize 24 kernels with a size of 3$\times$3 for each of the next three convolution layers. In the three FC layers, we employ 120, 84, and 24 or 10 neurons. To avoid overfitting, we apply dropout to the fully-connected layers with a probability of 0.5 similar to the original AlexNet. Figure \ref{fig:Alex} shows the AlexNet-inspired architecture.

\subsection{VGGNet}
Another well-known CNN used herein is VGGNet \cite{simonyan2014very}, 
which employs several 3$\times$3 filters instead of larger filters like AlexNet \cite{krizhevsky2012imagenet}. VGGNet, like AlexNet, uses ReLU activation functions in the network's hidden layers. Depending on the number of layers, there are multiple variations of VGG architecture. VGG-16, for example, includes 16 layers and about 138 million parameters. 
\textcolor{black}{Figure \ref{fig:VGG} depicts the structure of the VGG-inspired model developed in this work. We utilize only 9 layers consisting of 6 convolution layers, and 3 fully-connected layers.} 
We use 6 kernels for the first layer, 16 kernels for the second layer, 32 kernels for the third and fourth layers, and 48 kernels for the fifth and sixth layers. A non-overlapping max pooling layer with a 2$\times$2 filter size and strides of 2 follow each convolution block. For the fully-connected layers, we use the same number of neurons as AlexNet and LeNet models. 
\subsection{Model training}
The proposed models are trained using TensorFlow 2.6.2 \cite{tensorflow2015-whitepaper} with a categorical crossentropy loss function, a batch size of 128, and the Adam optimizer with a learning rate of 0.001 on the ASL Alphabet and ASL Digits datasets for 50 and 400 epochs, respectively. The intensities of the input images in both datasets are normalized from 0-255 to 0-1. Furthermore, data augmentation is 
utilized, which involves randomly rotating images in the range of 10 degrees, randomly shifting images horizontally and vertically by 10\%, and randomly zooming images by 10\%. During training, the best epoch is saved for each of the proposed models based on the lowest validation loss. Metrics like accuracy, precision, recall, and F1-score are used to evaluate the models.
\par

\section{SNN Conversion Toolbox Background}
\label{sec:snntoolboxsummary}

The SNN Conversion Toolbox proposed in \cite{snntoolbox} was designed with flexibility in mind to enable ANN models to be automatically converted to SNNs \cite{snntoolbox}.
During the SNN conversion process the toolbox performs several different steps: 

\begin{enumerate}
    \item Parse the trained ANN model along with the trained weights and activation values.
    \item Normalize, scale, and set the spiking neuron parameters such as the membrane potential thresholds, biases, and weights.
    \item Convert the layers of the ANN model into equivalent spiking representations using methods such as convolution unrolling.
    \item Deploy the resulting SNN on neuromorphic hardware or use an SNN simulator.
\end{enumerate}

The SNN Conversion Toolbox employs rate-based encoding which generates input spike trains with regular spiking frequencies over a duration parameter set in toolbox configuration settings. According to the SNN Conversion Toolbox documentation \cite{snntoolbox} this duration value correlates to the number of timesteps that each input is exposed to the SNN, in milliseconds, during inference. 
\textcolor{black}{While the duration parameter is noted to be measured in milliseconds, this time is not to be confused with the network's real-world latency. As we will see later in Section 6.2 the input duration and the latency are correlated in a linear fashion, but they are not equal in a 1-to-1 sense.}
The duration parameter does, however, affect the length of an input spike train to be proportional to the timesteps specified.
These spike trains then propagate through the network causing neurons in the SNN to fire if the neuron membrane potential is driven to or above its threshold. Then in the final layer, the classification layer, the neuron which fires the most is taken as the output class. These steps are a high-level view of the SNN Conversion Toolbox's ANN conversion methodology.
We refer the reader to \cite{snntoolbox} for detailed information on the methodologies used during the conversion process.


\begin{figure}[H]
         \centering
         \includegraphics[width=\textwidth]{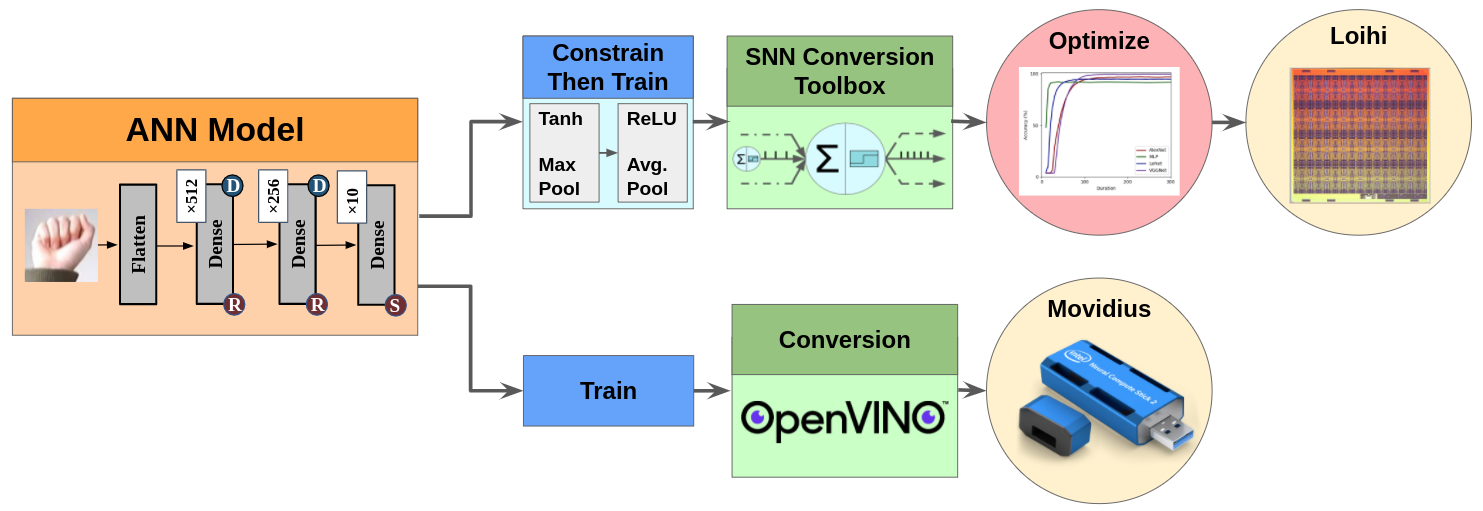}
    \caption{\textcolor{black}{Experimental methodology for ASL classification on Intel NCS2 and Loihi. ANN models are first created based off their deployement platform. For Loihi we constrain the networks by replacing \textit{TanH} and \textit{MaxPooling} with \textit{ReLU} and \textit{AveragePooling}, respectively, and then train the networks. Both sets of models are then converted to using respective tools for their platform. For Loihi we run the model multiple times with varying durations and optimize the duration parameter by minimizing the duration and maximizing accuracy. Finally, the models are deployed on their respective platforms.}}
     \label{fig:workflow}
\end{figure}

\section{Experiment Methodology}


Now that neuromorphic hardware, such as Intel's Loihi \cite{loihi_2018}, is becoming more readily available to researchers, SNNs can now be designed and tested on hardware specifically designed to accelerate SNNs. This allows SNNs to be fairly compared against their DNN counterparts on conventional hardware accelerators like GPUs and TPUs \cite{massa2020efficient, buettner2021heartbeat}.


\par

In this work, we deploy our developed ANN and SNN models on the Intel Neural Compute Stick 2 (NCS2) and the Intel Loihi neuromorphic platform, respectively. Our experimental methodology is depicted in Figure \ref{fig:workflow}. In particular, the Loihi experiments were performed on four Loihi chips for the accuracy measurements and a board of 32 Loihi chips called Nahuku-32 for power measurements. Each of the models in Figure \ref{fig:annarchs} were deployed on Loihi and NCS2 using the SNN Conversion Toolbox and OpenVINO APIs, respectively. From there, the performance of each hardware platform was measured with respect to latency, power, and accuracy.  



\subsection{Intel Neural Compute Stick 2}
The Intel Neural Compute Stick 2 (NCS2) is based on the Intel Movidius X Vision Processing Unit (VPU), which has 16 programmable SHAVE cores and a dedicated neural compute engine for hardware acceleration of deep neural network inferences. It features a base frequency of 700 MHz with a 16 nm technology node. Additionally, NCS2 has 4 GB memory with a maximum frequency of 1600 MHz. NCS2 supports 16-bit floating point operations, which we have found to be sufficient for our ASL classification tasks \cite{Intel:2019}.

\subsubsection{Conversion:}
To deploy the ANN models on NCS2, Intel provides a Python3 and C library called OpenVINO. Specifically, we used OpenVINO 2022.1.4 for our experiments. To deploy the models to the NCS2, the TensorFlow trained models were first frozen using TensorFlow's \textit{get\_concrete\_function()} and \textit{convert\_variables\_to\_constants\_v2()} functions. From these frozen models, the OpenVINO model optimizer was used to convert the models to a format that enables their deployment on the NCS2. Once the model is optimized for Intel NCS2, we then used the inference engine API built into OpenVINO to perform inference and subsequently measure the accuracy, latency, and power.

\subsubsection{Latency and Power Measurement:}
\label{sec:latency_power_measure}
Using OpenVINO, we performed 10 iterations over the entire test datasets, 7172 images for the ASL Alphabet dataset and 413 images for the ASL Digit dataset. The latencies were recorded over the 10 iterations for each image and then averaged at the end. 
We used a USB 3.0 measurement tool (MakerHawk UM34C \cite{WinNT}) along with a phone application to measure and record the power of the NCS2. This measurement device captures the overall system power usage, including USB I/O, due to its inline nature. The power measurement tool was attached to a computer's USB port along with the NCS2 and the idle and running power was recorded. The idle power was recorded for 5 minutes after plugging the NCS2 into the power measurement tool and the computer, and the voltage and current are measured every second. We then perform inference on the entire test dataset for 10 iterations and record the average running power. The difference between average running power and idle power is calculated as the \textit{inference power}.

\subsection{Intel Loihi}
Each experiment was performed on Intel's Neuromorphic Research Community's (INRC) cloud infrastructure for INRC members to test models on Intel's Loihi platform. In particular, two different nodes in the INRC cloud were used, one consisting of four Loihi chips and the other, Nahuku-32, consisting of 32 Loihi chips combined on a single board \cite{loihi_2021}. Each Loihi neuromorphic core or neuro-core can simulate up to 1024 spiking neurons or compartments, 4096 fan-out axons, 4096 fan-in axons, and 128 kilobytes of fan-in state memory \cite{loihi_2018}. The details of each Loihi platform node in the INRC cloud infrastructure are provided in Table \ref{tab:loihinodes}.

\begin{table}
    \centering
    \caption{Loihi Platforms.}
    \resizebox{\textwidth}{!}{
        \begin{tabular}{c c c c c c c c}
            \hline \hline
            Node & Chips & x86 Cores & Neurocores & Neurons & In-Axons & Out-Axons & State Memory\\
            \hline \\[-14pt] \hline
            Loihi & 4 & 12 & 512 & 524,288 & 2,097,152 & 2,097,152 & 65,536 KB\\
            Nahuku-32 & 32 & 96 & 4096 & 4,194,304 & 16,777,216 & 16,777,216 & 524,288 KB\\
            \hline \hline
        \end{tabular}
        }
    \label{tab:loihinodes}
\end{table}

\subsubsection{Conversion Methodology:}

In 2021, NxTF\cite{nxtf} was released enabling the SNN Conversion Toolbox \cite{snntoolbox} to convert a trained ANN model to an SNN and allow it to be deployed on Intel's Loihi platform. The SNN Conversion Toolbox, while supporting many of the common layers found in CNNs, does not support some ANN components and layers on Loihi. To combat this, we employ a constrain-then-train method to ensure compatibility with the Loihi backend for the SNN Conversion Toolbox before conversion. 
The constrain-then-train method consists of first replacing the max-pooling layers with average-pooling layers and then changing the activation functions from $TanH$ to $ReLU$. We then train the models with the same configuration as before and then input the trained models into the SNN Conversion Toolbox.

\begin{table}[H]
\color{black}
\caption{\textcolor{black}{Loihi Neuron Parameters}}
\label{tab:loihi_neuron_parameters}
\resizebox{\textwidth}{!}{
\begin{tabular}{lcl}
\hline
\multicolumn{1}{c}{\textbf{Parameter}} & \textbf{Value} & \multicolumn{1}{c}{\textbf{Description}} \\ \hline
Reset Mode* & soft & Resets membrane potential to u(t) = u(t-1) - vTh. \\ \hline
Threshold-Input Ratio* & 1 & Balances vTh w.r.t. input voltage change. \\ \hline
Compart. Bias Exponent* & 6 & bias = \textless{}biasMant\textgreater{}*2\textasciicircum{}\textless{}biasExp\textgreater{} \\ \hline
Voltage Threshold Mantissa* & 512 & Used to compute voltage threshold vTh = \textless{}vThMant\textgreater{}*2\textasciicircum{}6 \\ \hline
Weight Bit Precision* & 8 & Bit precision to represent weight connections. \\ \hline
Bias Bit Precision* & 12 & Bit precision to represent bias connections. \\ \hline
Compart. Voltage Decay & 0 & Voltage decay constant per timestep. \\ \hline
Compart. Current Decay & 4095 & Current decay constant per timestep. \\ \hline
Refractory Delay & 1 & Enables refractory period after compartment spiking. \\ \hline
Functional State & 2 & Idle compartment behavior waiting for excitatory stimulus. \\ \hline
\end{tabular}
}
\begin{threeparttable}
\begin{tablenotes}
 \item [*] \footnotesize Configurable parameters through the SNN Conversion Toolbox.
\end{tablenotes}
\end{threeparttable}
\end{table}

\textcolor{black}{Before converting the ANNs into SNNs, the SNN Conversion Toolbox requires settings and parameters to be initialized in the form of a configuration file. This configuration file configures settings like what simulator to use, the input duration in milliseconds, and the neuron parameters. In most cases, we use the default parameters specified in the SNN Conversion Toolbox examples. However, particular attention was given to the duration and the Loihi neuron parameters.
In \cite{loihi_2018, loihi_2018_mapping, loihi_2021}, Loihi neurons are described as a variant of the current-based (CUBA) leaky-integrate-and-fire (LIF) neurons described in \cite{cuba, neuronal_dynamics}. The specific Loihi compartment/neuron parameters used in our experiments are provided and described in Table \ref{tab:loihi_neuron_parameters}. For more specific information on CUBA LIF neurons and Loihi, we refer the reader to \cite{cuba, neuronal_dynamics} and \cite{loihi_2018, loihi_2018_mapping, loihi_2021}.}

As mentioned in Section \ref{sec:snntoolboxsummary}, after the trained ANN is converted to an SNN, the SNN Conversion Toolbox uses the SNN parameters and layer information to deploy the models on neuromorphic hardware or a software simulator. In this case, the toolbox uses the information about the converted SNN structure to appropriately partition the Loihi neuro-cores. This partitioning process is described in \cite{nxtf} and consists of optimization techniques that ensure that the neuro-cores can communicate efficiently.
The specific number of neuro-cores which were partitioned for each layer of our converted SNN models can be seen in Tables \ref{tab:params_parts_alpha} and \ref{tab:params_parts_digits}. Once the partitioning process has succeeded, the SNN Conversion Toolbox proceeds to deploy the SNN on Loihi and begins performing inferences on the test dataset.

\begin{table}[]
\centering
\renewcommand{\arraystretch}{1.2}
\caption{ASL Alphabet Model Parameters and Loihi Core Partitioning}
\resizebox{\textwidth}{!}{
\begin{tabular}{ccccccccccccc}
\hline \hline
\multirow{2}{*}{Layer} &  & \multicolumn{2}{c}{MLP}                                        &  & \multicolumn{2}{c}{LeNet}                                     &  & \multicolumn{2}{c}{AlexNet}                                   &  & \multicolumn{2}{c}{VGGNet}                                    \\ \cline{3-4} \cline{6-7} \cline{9-10} \cline{12-13} 
                       &  & Param  & \begin{tabular}[c]{@{}c@{}}Loihi \vspace{-0.3cm}\\ Cores\end{tabular} &  & Param & \begin{tabular}[c]{@{}c@{}}Loihi\vspace{-0.3cm}\\ Cores\end{tabular} &  & Param & \begin{tabular}[c]{@{}c@{}}Loihi\vspace{-0.3cm}\\ Cores\end{tabular} &  & Param & \begin{tabular}[c]{@{}c@{}}Loihi\vspace{-0.3cm}\\ Cores\end{tabular} \\ \hline \\[-22pt] \hline
Conv1                  &  & -       & -           &  & 156    & 9           &  & 156     & 12           &  & 60     & 12   \\
Conv2                  &  & -       & -           &  & 2416   & 4           &  & 1812    & 7           &  & 880    & 7     \\
Conv3                  &  & -       & -           &  & -      & -           &  & 2616    & 3           &  & 4640   & 4      \\
Conv4                  &  & -       & -           &  & -      & -           &  & 5208    & 4           &  & 9248   & 7      \\
Conv5                  &  & -       & -           &  & -      & -           &  & 5208    & 4           &  & 13872  & 2      \\
Conv6                  &  & -       & -           &  & -      & -           &  & -       & -           &  & 20784  & 3      \\
FC1                    &  & 401920  & 10           &  & 30849  & 1           &  & 26040   & 1           &  & 5880   & 1     \\
FC2                    &  & 131328  & 4           &  & 10164  & 1           &  & 10164   & 1           &  & 10164  & 1      \\
Output                 &  & 6168    & 1           &  & 2040   & 1           &  & 2040    & 1           &  & 2040   & 1      \\ \hline
\textbf{Total}                  &  & 539,416 & 18           &  & 45,616 & 21           &  & 53,244  & 41           &  & 67,568 & 47 \\ \hline \hline
\end{tabular}}
\label{tab:params_parts_alpha}
\end{table}

\begin{table}[]
\centering
\renewcommand{\arraystretch}{1.2}
\caption{ASL Digits Model Parameters and Loihi Core Partitioning}
\resizebox{\textwidth}{!}{
\begin{tabular}{ccccccccccccc}
\hline \hline
\multirow{2}{*}{Layer} &  & \multicolumn{2}{c}{MLP}                                        &  & \multicolumn{2}{c}{LeNet}                                     &  & \multicolumn{2}{c}{AlexNet}                                   &  & \multicolumn{2}{c}{VGGNet}                                    \\ \cline{3-4} \cline{6-7} \cline{9-10} \cline{12-13} 
                       &  & Param  & \begin{tabular}[c]{@{}c@{}}Loihi\vspace{-0.3cm}\\ Cores\end{tabular} &  & Param & \begin{tabular}[c]{@{}c@{}}Loihi\vspace{-0.3cm}\\ Cores\end{tabular} &  & Param & \begin{tabular}[c]{@{}c@{}}Loihi\vspace{-0.3cm}\\ Cores\end{tabular} &  & Param & \begin{tabular}[c]{@{}c@{}}Loihi\vspace{-0.3cm}\\ Cores\end{tabular} \\ \hline \\[-22pt] \hline
Output &  & 2570    & 1 &  & 850    & 1 &  & 850    & 1 &  & 850    & 1 \\ \hline
\textbf{Total}   &  & 535,818 & 18 &  & 44,426 & 21 &  & 52,054 & 41 &  & 66,378 & 47\\ \hline \hline
\end{tabular}}
\label{tab:params_parts_digits}
\end{table}

\subsubsection{Hardware Independent Simulation}
\textcolor{black}{To gauge the energy efficiency of each network independent of the neuromorphic hardware platform, we simulated the converted SNNs using the INIsim SNN simulator built into the SNN Conversion Toolbox. In Table \ref{table:afr}, we report the average firing rate (AFR) of the network as a whole. This average firing rate is computed by}
\begin{equation}
    \color{black}
    AFR = \frac{N_{spikes}}{Batch*M_{neurons}*Duration}
\end{equation}
\textcolor{black}{where $N_{spikes}$ is the number of spikes generated during the simulation for each inference, $Batch$ is the batch size of the input which we set to 1, $M_{neurons}$ is the number of neurons in the network, and $Duration$ is the simulation duration parameter which we set to 500 ms to obtain an upper bound on the AFR.}

\begin{table}[H]
\color{black}
\caption{\textcolor{black}{Average Network Firing Rate}}
\label{table:afr}
\centering
\begin{tabular}{ccc}
\hline
\multicolumn{3}{c}{\textbf{Average Network Firing Rates}} \\ \hline
\textbf{Model} & \textbf{ASL Alphabet} & \textbf{ASL Digits} \\ \hline
MLP & 0.050 & 0.039 \\ \hline
LeNet & 0.358 & 0.313 \\ \hline
AlexNet & 0.098 & 0.182 \\ \hline
VGGNet & 0.062 & 0.039 \\ \hline
\end{tabular}
\end{table}

\subsubsection{Latency and Power Measurement:}
\label{sec:lat_power_loihi}

After designing the ANNs using the aforementioned constrain-then-train method and converting them to SNNs, we performed multiple experiments with varying durations, a hyper-parameter defined in Section \ref{sec:snntoolboxsummary}. \textcolor{black}{The duration point space was then searched to achieve one of two objectives: (1) attain high accuracy without duration considerations or (2) achieve a balance between duration and accuracy.} To collect the latency and power during inference, we set the \textit{profile\_performance} configuration setting in the toolbox while targeting the Nahuku-32 board. Nahuku-32 was chosen since it contains the necessary power and latency recording hardware \cite{loihi_2021}.
\textcolor{black}{While the Nahuku-32 board contains 32 Loihi chips in a single server node, according to our experiments all of the models utilize one Loihi chip and less than the 128 neuro-cores available as seen in Tables \ref{tab:params_parts_alpha} and \ref{tab:params_parts_digits}.}
For each of the varying durations, the latency and power metrics were recorded when the SNN models were deployed on the Nahuku-32 board. After completing all inferences, the toolbox then reports the power usage for the neuro-cores, x86 cores, and the total system power. For each model, power measurements for various durations were averaged together to attain a typical power usage of the model on Loihi regardless of the duration.



\vspace{-2mm}
\section{Results}
\label{sec:results}
In this section, we compare the SNNs' and ANNs' performance in terms of accuracy, latency, power, and energy consumption. We also provide insights into the techniques utilized to obtain high SNN accuracy and then provide an analysis on the balance between accuracy and latency. Finally, the SNN deployed on Loihi is compared to the ANN deployed on Intel NCS2 in terms of latency, power, and energy.

\subsection{Accuracy Analysis}
\label{sec:acc_analysis}
SNNs have been shown in previous works to have comparable if not better accuracy than ANNs due to the introduction of noise by the nature of spikes approximating the floating point values in an ANN \cite{buettner2021heartbeat,blouw2019benchmarking}. \textcolor{black}{Table \ref{table:accuracies} shows the average accuracy and standard deviation values obtained for the ANN, constrained ANN (C-ANN), and SNN models for the ASL Alphabet and ASL Digits over ten different trials. The confidence intervals for all models and datasets are calculated in Appendix D, which shows a narrow margin of uncertainty.}

\begin{table}[H]
\centering
\color{black}
\caption{\textcolor{black}{Average ASL Accuracy \& Standard Deviation}}
\begin{tabular}{cccccccc}
\hline
\multirow{2}{*}{\textbf{Dataset}} & \multirow{2}{*}{\textbf{Architecture}} & \multicolumn{3}{c}{\textbf{Accuracy (\%)}} & \multicolumn{3}{c}{\textbf{Standard Deviation (\%)}} \\ \cline{3-8} 
 &  & ANN & C-ANN & SNN & ANN & C-ANN & SNN \\ \hline
\multirow{4}{*}{\begin{tabular}[c]{@{}c@{}}ASL\\ Alphabet\end{tabular}} & MLP & 92.28 & - & 91.18 & 0.88 & - & 1.28 \\
 & LeNet & 98.30 & 97.71 & 94.82 & 0.43 & 0.91 & 1.49 \\
 & AlexNet & 98.99 & 99.02 & 97.50 & 0.68 & 0.53 & 0.78 \\
 & VGGNet & \textbf{99.38} & \textbf{99.13} & \textbf{98.82} & 0.31 & 0.42 & 0.56 \\ \hline
\multirow{4}{*}{\begin{tabular}[c]{@{}c@{}}ASL\\ Digits\end{tabular}} & MLP & 86.61 & - & 86.05 & 1.06 & - & 1.28 \\
 & LeNet & 95.04 & 96.44 & 95.78 & 0.46 & 1.03 & 1.12 \\
 & AlexNet & 98.77 & 98.60 & 97.01 & 0.62 & 0.39 & 1.37 \\
 & VGGNet & \textbf{99.20} & \textbf{99.08} & \textbf{97.60} & 0.38 & 0.37 & 1.21 \\ \hline
 \label{table:accuracies}
\end{tabular}
\end{table}

First, we compare the ANN versus C-ANN model accuracies before conversion. As listed in Table \ref{table:accuracies}, the VGGNet ANN models achieve the  highest values for accuracy. In particular, for the ASL Alphabet, the VGG ANN realizes the best average accuracy of \textcolor{black}{99.38\%} and standard deviation of \textcolor{black}{0.31\%}, while the ASL Digit's VGG ANN has the best accuracy and standard deviation of \textcolor{black}{99.20\%} and \textcolor{black}{0.38\%} respectively. For both datasets, the ASL Alphabet and Digits, the MLP networks performed the worst with the lowest accuracies. Since the MLP models did not contain any $TanH$ activations or pooling layers, no constrained models were created, and thus, the C-ANN metrics have been left empty. We can also see that the AlexNet C-ANN on the ASL Alphabet performs marginally better when compared to the conventional ANN with a \textcolor{black}{0.03\%} difference in accuracy and a \textcolor{black}{0.15\%} difference in their standard deviation. 
Table \ref{table:accuracies} also shows that the average accuracy of the ASL Digits LeNet C-ANN is \textcolor{black}{1.4\%} higher than that of its ANN, but the standard deviation of the C-ANN is \textcolor{black}{0.57\%} higher than the ANN's.


In Table \ref{table:accuracies}, we also present the average accuracies and standard deviation obtained from the deployed SNN models run on the durations which maximize accuracy. For ASL Alphabet, the VGGNet SNN achieves the best SNN average accuracy of \textcolor{black}{98.82\%} and standard deviation of \textcolor{black}{0.56\%} compared to the other SNN architectures. This also holds true for the ASL Digits dataset where the highest SNN average accuracy and standard deviation is \textcolor{black}{97.60\%} and \textcolor{black}{1.21\%}, respectively. Comparing the VGGNet ANN and SNN models for the ASL Alphabet, the SNN loses just \textcolor{black}{0.56\%} of its accuracy after conversion. Similarly, the ASL Digit VGG SNN loses \textcolor{black}{1.60\%} compared to VGG ANN. A similar pattern can be seen for the other models in both datasets where SNN accuracies are lower than the ANNs except the LeNet model on ASL digits. In this case, LeNet outperforms the ANN with a difference of \textcolor{black}{0.74\%}, but the C-ANN still surpasses the SNN with a difference of \textcolor{black}{0.66\%}.  
\textcolor{black}{In Table \ref{tab:acc_comparion}, we compare our best accuracy results to those of the previous works. We can see that, in comparison to the other works, our ANN and SNN models achieve comparable accuracy with significantly smaller networks in terms of network parameters.
}

\begin{table}[h]
\centering
\color{black}
\caption{\textcolor{black}{Previous Work Model Comparison with Best Accuracies}}
\label{tab:acc_comparion}
\begin{tabular}{clcr}
\hline
\textbf{Datset} & \textbf{Paper} & \textbf{Accuracy} & \multicolumn{1}{c}{\textbf{Parameters}} \\ \hline
\multirow{5}{*}{ASL-Alphabet} & A Mannan et. al.  \cite{mannan2022hypertuned} & 99.97 & 2,994,649 \\
 & C Can et. al. \cite{can2021deep} & 99.91 & 1,355,722 \\
 & J Fregoso et. al. \cite{fregoso2021optimization} & \textbf{99.98} & - \\
 & Our ANN & 99.93 & \textbf{67,568} \\
 & Our SNN & 99.30 & 82,150 \\ \hline
\multirow{4}{*}{ASL-Digits} & MM Rahman et. al.  \cite{rahman2019new} & \textbf{99.99} & - \\
 & H Xiao et. al. \cite{xiao2022sign} & 99.52 & - \\
 & Our ANN & 99.76 & \textbf{66,378} \\
 & Our SNN & 99.03 & 80,960 \\ \hline
\end{tabular}
\end{table}

\textcolor{black}{In Table \ref{tab:acc_comparion} we compare our accuracy results to previous works. To compare our work with other cutting-edge models, we selected the experiment with the best accuracy out of 10 distinct trials.We can see that, in comparison to the other works, the accuracies for both our ANNs and SNNs are rather close to the state-of-the-art models which use the same datasets. Our ASL-Alphabet SNN is just 0.68\% less accurate than the best ANN and our ASL-Digits SNN is only 0.96\% less accurate than the best ANN for its dataset. Table \ref{tab:acc_comparion} also shows that our ANN and SNN models both realize high accuracy with much smaller networks in terms of network parameters.}

\subsection{Latency}
\begin{figure}[H]
    \centering
    \subfloat[]{
        \includegraphics[width=.45\textwidth]{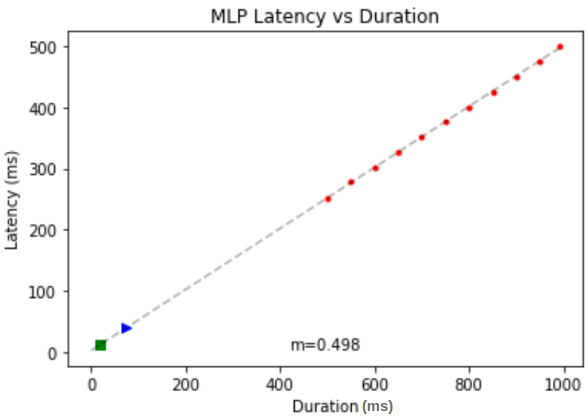}
    }
        \subfloat[]{
        \includegraphics[width=.45\textwidth]{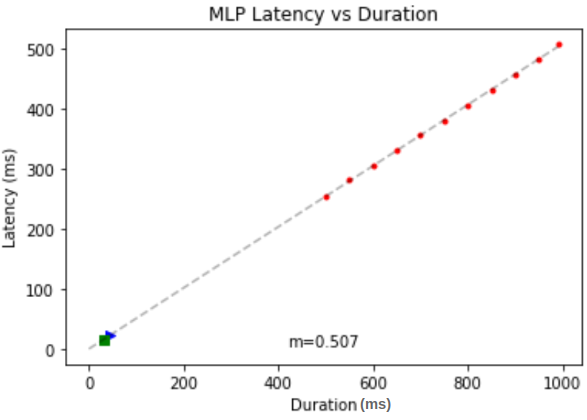}
    }
    \caption{MLP Latency vs. Duration graphs for (a) ASL Alphabet and (b) ASL Digit dataset. See Appendix \ref{sec:duration_latency_lines} for the other models. Legend: \shapelabel{circle}{red}{} test points, \shapelabel{regular polygon,regular polygon sides=3}{blue}{} best accuracy, \shapelabel{rectangle}{black!30!green}{} balanced duration-accuracy point.}
    \label{fig:mlp_duration_latency}
\end{figure}

As mentioned in Section \ref{sec:snntoolboxsummary}, the duration parameter is the number of timesteps, in milliseconds, that each input is exposed to the SNN. Thus, modifying this duration value should have a linear effect on the inference latency. As mentioned in Section \ref{sec:lat_power_loihi}, the Nahuku-32 board does contain the power and latency measurement hardware, but the polling is limited to a 30-40ms time resolution \cite{ncl_models}. 
This limit, thus, prevents the power and latency measurements when running on lower values of duration. To mitigate the polling issue, we performed our experiments using a set of higher and uniformly distributed values of duration to drive the time for inference above the 30-40ms threshold. For each of the higher durations, the models are run and the latency is recorded and plotted seen as red circles \shapelabel{circle}{red}{} on the graphs in Figure \ref{fig:mlp_duration_latency} (and the figures in \ref{sec:duration_latency_lines}). From there, the least squares method was used to generate the line of best fit. As seen in the graphs of Figure \ref{fig:mlp_duration_latency} (and in \ref{sec:duration_latency_lines}), the relationship between duration and latency is roughly linear for all the models on higher durations. Thus, from here on, we use the fitted line to obtain the SNN inference latency for various duration values. 


\subsubsection{Loihi Accuracy and Duration/Latency Balance:}
\begin{figure}[H]
     \begin{subfigure}[b]{0.45\textwidth}
         \centering
         \includegraphics[width=\textwidth]{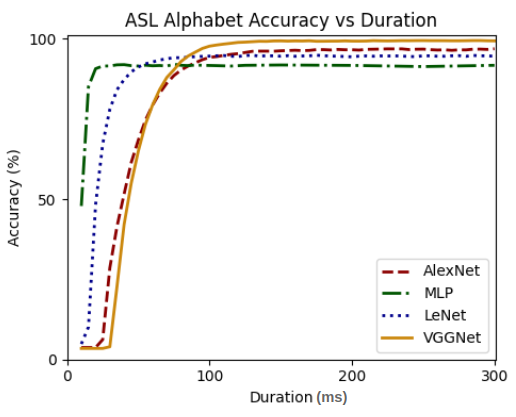}
         \caption{\textit{ASL Alphabet}}
     \end{subfigure}
     \begin{subfigure}[b]{0.45\textwidth}
         \centering
         \includegraphics[width=\textwidth]{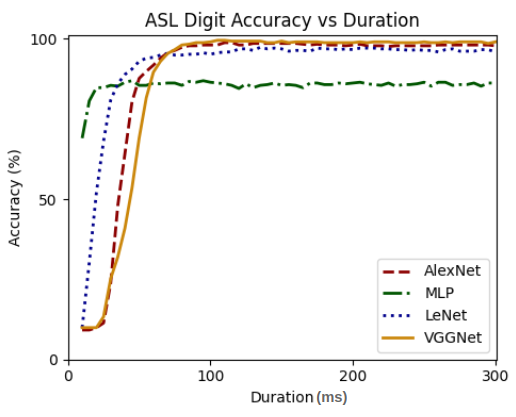}
         \caption{\textit{ASL Digits}}
     \end{subfigure}
     \caption{Accuracy-Duration Tradeoff. The SNN model's accuracy on test dateset is related to the number of timesteps in which each sample is given.}
     \label{fig:tradeoff}
\end{figure}

As shown in Figure \ref{fig:tradeoff}, as the duration/latency increases so does the accuracy for various SNN models run on Loihi.However, after a specific duration, the accuracy plateaus and there is little to no gain in accuracy at the cost of significantly increasing duration and therefore latency. Thus, we aim to achieve a balance between the accuracy and duration to attain acceptable accuracy while reducing the latency.

The blue triangles \shapelabel{regular polygon,regular polygon sides=3}{blue}{} on the graphs in Figure \ref{fig:mlp_duration_latency} (and the figures in \ref{sec:duration_latency_lines}) represent the maximum possible accuracy for the respective models between 5 ms and 300 ms with increments of 5 ms. The quantitative values of the blue point can be found as the ``Best Accuracy'' metrics in Table \ref{table:relaxed_accuracy}.
While it is possible that a higher accuracy could be achieved after 300 duration timesteps, this accuracy would incur a significantly higher inference latency.

Table \ref{table:relaxed_accuracy} demonstrates the accuracy and duration trade-offs, for multiple relaxed accuracy values ranging from 1.0\% to 5.0\% accuracy reductions. As seen in Table \ref{table:relaxed_accuracy} there are significant reductions in duration if a small reduction in accuracy is allowed.

\begin{table}[H]
\caption{Relaxed accuracy yielding lower duration.}
\label{table:relaxed_accuracy}
\centering
\resizebox{\textwidth}{!}{
\begin{tabular}{ccccccccc}
\hline \hline
\multicolumn{1}{c}{\multirow{2}{*}{Dataset}} & \multicolumn{1}{c}{\multirow{2}{*}{SNN Model}} &
Best &
\multicolumn{6}{c}{Relaxed Accuracy Range Durations (ms)} \\ \cline{4-9}
\multicolumn{1}{c}{} & \multicolumn{1}{c}{} & Accuracy & 0.0\% & \textless{}1.0\% & \textless{}2.0\% & \textless{}3.0\% & \textless{}4.0\% & \textless{}5.0\% \\
\hline \\[-14pt] \hline
\multirow{4}{*}{Alphabet} 
 & MLP & 92.07\% & 75 & 25 & 20 & 20 & 20 & 20 \\
 & LeNet & 94.87\% & 130 & 75 & 65 & 55 & 50 & 50 \\
 & AlexNet & 96.88\% & 300 & 130 & 115 & 100 & 95 & 90 \\
 & VGGNet & 99.44\% & 250 & 115 & 100 & 95 & 90 & 85 \\
\hline
\multirow{4}{*}{Digits}
 & MLP & 86.92\% & 45 & 40 & 30 & 20 & 20 & 20 \\
 & LeNet & 97.34\% & 135 & 120 & 70 & 65 & 55 & 50 \\
 & AlexNet & 98.79\% & 110 & 85 & 80 & 75 & 70 & 65 \\
 & VGGNet & 99.52\% & 105 & 90 & 80 & 75 & 75 & 70 \\
 \hline \hline
\end{tabular}
}
\end{table}

In Table \ref{table:trade_off_alphabet} and \ref{table:trade_off_digits}, we use the duration point from Table \ref{table:relaxed_accuracy} where the accuracy drop is no more than 2.0\%. This threshold was chosen because it reduces the duration, therefore the latency, without significantly reducing the accuracy. However, another accuracy threshold can be used depending on the application's sensitivity to accuracy or latency. The green squares \shapelabel{regular polygon,regular polygon sides=4}{black!30!green}{} on the graphs in Figure \ref{fig:mlp_duration_latency} (and the graphs in \ref{sec:duration_latency_lines}) exhibit the 2.0\% relaxed accuracy point. The accuracy, duration, and latency values for this point are provided in Tables \ref{table:trade_off_alphabet} and \ref{table:trade_off_digits} as the ``Balanced Point'' metrics.

\begin{table}[H]
\centering
\caption{ASL Alphabet SNN best accuracy metrics versus metrics for balanced accuracy-duration on Loihi.}
\label{table:trade_off_alphabet}
\resizebox{\textwidth}{!}{
\begin{tabular}{clccclccc}
\hline \hline
\multirow{2}{*}{SNN Model} &  & \multicolumn{3}{c}{Best Point \shapelabel{regular polygon,regular polygon sides=3}{blue}{}} &  & \multicolumn{3}{c}{Balanced Point \shapelabel{regular polygon,regular polygon sides=4}{black!30!green}{}} \\ \cline{3-5} \cline{7-9} 
 &  & Duration (ms) & Latency (ms) & Accuracy (\%) &  & Duration (ms) & Latency (ms) & Accuracy (\%) \\ \hline \\[-14pt] \hline
MLP &  & 75 & 40.06 & 92.07 &  & 20 & 12.64 & 90.67 \\
LeNet &  & 130 & 13.21 & 94.87 &  & 65 & 8.97 & 93.39 \\
AlexNet &  & 300 & 20.69 & 96.88 &  & 115 & 11.40 & 95.13 \\
VGGNet &  & 250 & 14.03 & 99.44 &  & 100 & 8.03 & 97.70 \\ \hline \hline
\end{tabular}}
\end{table}

\begin{table}[H]
\centering
\caption{ASL Digit SNN best accuracy metrics versus metrics for balanced accuracy-duration on Loihi.}
\label{table:trade_off_digits}
\resizebox{\textwidth}{!}{
\begin{tabular}{clccclccc}
\hline \hline
\multirow{2}{*}{SNN Model} &  & \multicolumn{3}{c}{Best Point \shapelabel{regular polygon,regular polygon sides=3}{blue}{}} &  & \multicolumn{3}{c}{Balanced Point \shapelabel{regular polygon,regular polygon sides=4}{black!30!green}{}} \\ \cline{3-5} \cline{7-9} 
 &  & Duration (ms) & Latency (ms) & Accuracy (\%) &  & Duration (ms) & Latency (ms) & Accuracy (\%) \\ \hline \\ [-14pt] \hline
MLP &  & 45 & 23.89 & 86.92 &  & 30 & 16.29 & 85.47 \\
LeNet &  & 135 & 13.56 & 97.34 &  & 70 & 9.67 & 95.40 \\
AlexNet &  & 110 & 10.67 & 98.79 &  & 80 & 9.04 & 97.34 \\
VGGNet &  & 105 & 8.72 & 99.52 &  & 80 & 7.77 & 98.06 \\ \hline \hline
\end{tabular}
}
\end{table}

As shown in Tables \ref{table:trade_off_alphabet} and \ref{table:trade_off_digits}, the balanced point's latency is significantly reduced compared to that of the best accuracy point. 
The ASL Alphabet appears to benefit the most from the accuracy-duration balancing. As seen in Table \ref{table:trade_off_alphabet}, the MLP's SNN accuracy drops just 1.4\% but the duration decreases by 73.33\% equating to a 68.44\% drop in latency. VGGNet's accuracy is reduced by 1.74\% after balancing, from 99.44\% to 97.70\%, and the duration is subsequently reduced by 60.0\% and latency by 42.77\%. The accuracy loss for the other two models, LeNet and AlexNet, is 1.48\% and 1.75\%, respectively. Their duration reduction is 50.0\% and 61.67\% equating to a latency reduction of 32.10\% and 44.90\%, respectively.

For the ASL Digits dataset, the MLP SNN has the most significant difference in duration and latency with a 31.81\% reduction in latency and a 1.45\% drop in accuracy. The duration reductions for LeNet, AlexNet, and VGGNet are 48.15\%, 27.27\%, and 23.81\%, while the accuracy loss is 1.94\%, 1.45\%, and 1.46\%, respectively. The latency reductions for the three SNN models are thus 28.69\%, 15.28\%, and 10.89\%, respectively. Hence, for applications that can tolerate more error, it is evident that lower latency configurations can be employed.

\subsection{Intel Loihi and NCS2 Compared}  
Here, we compare the performance of ANNs on Intel's NCS2 with the performance of SNNs on Intel's Loihi neuromorphic chips in terms of latency, power, and energy.


\begin{table}[H]
\caption{Movidius vs Loihi power comparison on ASL Alphabet}
\renewcommand{\arraystretch}{1.1}
\resizebox{\columnwidth}{!}{
\begin{tabular}{clcclccccccc}
\hline \hline
\multirow{3}{*}{Model} &  & \multicolumn{2}{c}{NCS2 Power (mW)}            &  & \multicolumn{7}{c}{Loihi Power (mW)}                                                   \\ \cline{3-4} \cline{6-12} 
                       &  & \multirow{2}{*}{Idle} & \multirow{2}{*}{Running} &  & \multicolumn{3}{c}{X86 Cores} & \multicolumn{1}{l}{} & \multicolumn{3}{c}{Neuro-cores} \\ \cline{6-8} \cline{10-12} 
                       &  &                         &                          &  & Static  & Dynamic    & Total  & \multicolumn{1}{l}{} & Static   & Dynamic   & Total    \\ \hline \\[-15.5pt] \hline 
MLP                    &  & 635                     & 1513                      &  & 0.254   & 18.180     & 18.434 &                      & 15.440   & 27.500      & 42.940    \\
Lenet                  &  & 635                     & 1487                      &  & 0.255   & 18.736     & 18.991 &                      & 18.119   & 20.954    & 39.073   \\
AlexNet                &  & 635                     & 1482                      &  & 0.263   & 18.806     & 19.069 &                      & 36.423   & 23.595    & 60.018   \\
VGGNet                &  & 635                     & 1460
&  & 0.253   & 19.566 & 19.819  &                      & 40.180    & 22.976    & 63.156   \\ \hline \hline
\end{tabular}}
\label{table:power_comp_Alphabet}
\end{table}

\begin{table}[H]
\caption{Movidius vs Loihi power comparison on ASL Digits}
\renewcommand{\arraystretch}{1.1}
\resizebox{\columnwidth}{!}{
\begin{tabular}{clcclccccccc}
\hline \hline
\multirow{3}{*}{Model} &  & \multicolumn{2}{c}{NCS2 Power (mW)} &  & \multicolumn{7}{c}{Loihi Power (mW)} \\ \cline{3-4} \cline{6-12} 
&  & \multirow{2}{*}{Idle} & \multirow{2}{*}{Running} &  & \multicolumn{3}{c}{X86 Cores} & \multicolumn{1}{l}{} & \multicolumn{3}{c}{Neuro-cores} \\ \cline{6-8} \cline{10-12} 
& & & & & Static & Dynamic & Total   & \multicolumn{1}{l}{} & Static & Dynamic & Total \\ \hline \\[-15.5pt] \hline
MLP & & 635 & 1484 & & 0.255 & 18.187 & 18.442 & & 15.476 & 26.312 & 41.788   \\
Lenet & & 635 & 1473 & & 0.262 & 18.835 & 19.097 & & 18.602   & 22.831 & 41.433   \\
AlexNet & & 635 & 1472 & & 0.258 & 18.848 & 19.106 & & 35.699 & 35.535 & 71.234 \\
VGGNet & & 635 & 1475 & & 0.267 & 18.964 & 19.231 & & 42.423 & 22.890 & 65.313 \\ \hline \hline
\end{tabular}}
\label{table:power_comp_Digits}
\end{table}

To measure the inference power consumption on Loihi, the accuracy-latency balanced SNN models discussed in the previous subsection are evaluated on the Nahuku-32 board. The LeNet SNN, for the ASL Alphabet, and MLP SNN, for the ASL Digits, have the lowest inference power with 58.06 mW and 60.23 mW, respectively, as seen in Tables \ref{table:Comprehensive_Alphabets} and \ref{table:Comprehensive_Digits}. 
\textcolor{black}{To analyze the effect the number of parameters in a network has on the power, latency, and energy we also ran the ASL Alphabet LeNet SNN model using varying numbers of kernels in the convolution layers to increase/decrease the total number of parameters without changing the number of layers. As seen in Table \ref{table:lenet_varying_params} in Appendix C, increasing the number of parameters in a network also increases the power of a network showing that power consumption is not just dependent on model depth.} 
The AlexNet SNN, on the other hand, has the highest inference power, with the ASL Alphabet and ASL Digits inference powers of 79.09 mW and 90.34 mW, respectively (see Tables \ref{table:power_comp_Alphabet} and \ref{table:power_comp_Digits}). For NCS2, an idle power of 635 mW is measured using the method discussed in Section \ref{sec:latency_power_measure}. The MLP network has somewhat higher inference power for both datasets compared to all other models running on NCS2, with 878 mW and 849 mW for the ASL Alphabet and ASL Digits, respectively. On the other hand, VGGNet has the lowest inference power on the ASL Alphabet, with an inference power of 825 mW. On ASL Digits, however, the CNN architectures have very comparable inference power. 
In terms of latency, the MLP on the NCS2 has the least latency for both ASL Alphabet and ASL digits, with 2.13 ms and 2.06 ms values, respectively, whereas VGGNet has the greatest latency, with values of 2.59 ms and 2.47 ms, respectively. 
On both datasets, on Loihi, the VGGNet SNN had the least inference latency. The SNN models running on Loihi, in general, have a greater inference latency, ranging from 8.03 ms to 12.64 ms for the ASL alphabet and 7.77 ms to 16.29 ms for the ASL digits. 
\textcolor{black}{This higher latency, when compared to NCS2, can be attributed to the rate-based spike coding used by the SNN Conversion Toolbox as well as the inclusion of three x86 cores in the current implementation of Loihi \cite{loihi_2021, buettner2021heartbeat}.
In previous works \cite{snn_coding_2021}, rate-based encoding has been shown to significantly increase latency due to features being encoded into the frequency of spikes in spike trains. These spike trains must then be sufficiently long enough with a sufficient firing rate to propagate through the network to allow the network to classify/learn. This can be seen in Figure 6 showing a network's accuracy being dependent on the input duration i.e. the input spike train length. Moreover, in \cite{buettner2021heartbeat}, the authors show that most of the higher latency is attributed to I/O-related tasks such as transferring data to and from synchronous and asynchronous domains as well as between the host CPU and devices.
This supports the claims from Intel, in \cite{loihi_2021}, that the current implementation's x86 cores are sub-optimal and can introduce latency overhead due to the communication differences between synchronous computations on the x86 cores and asynchronous computations on the neuro-cores \cite{loihi_2021}.}



In terms of inference energy consumption, the LeNet SNN model on Loihi has the lowest inference energy, with 0.52 mJ and 0.59 mJ for the ASL Alphabet and ASL Digits, respectively. On the ASL Alphabet images, the VGGNet SNN model on Loihi consumes only 0.06 mJ more energy than LeNet but achieves 4.31\% higher accuracy. VGGNet also consumes 0.07 mJ more inference energy for the ASL Digits, although it is 2.66\% more accurate compared to the LeNet SNN. As a result, the VGGNet SNN is the more appropriate choice for applications that require a model with high accuracy and low energy consumption.

A comprehensive comparison of the SNN models implemented on Loihi and ANN models deployed on NCS2 is provided in Tables \ref{table:Comprehensive_Alphabets} and \ref{table:Comprehensive_Digits}. All of the SNN models on Loihi use less power and energy compared to ANN models. On the ASL Alphabet, for example, while VGGNet loses just 1.9\% accuracy after converting from the ANN to an SNN, it consumes 11.48$\times$ less inference power and 3.69$\times$ less inference energy. For the ASL Digits dataset, the VGGNet SNN running on Loihi consumes 9.94$\times$ less inference power and 3.15$\times$ lower energy compared to VGGNet ANN on NCS2. 



\begin{table}[H]
\caption{Movidious vs Loihi - Comprehensive Analysis - ASL Alphabet}
\centering
\renewcommand{\arraystretch}{1.2}
\resizebox{\columnwidth}{!}{%
\begin{tabular}{clcccclcccc}
\hline \hline
\multirow{2}{*}{\begin{tabular}[c]{@{}c@{}}Benchmarking\\ Metric\end{tabular}} &  & \multicolumn{4}{c}{NCS2}     &  & \multicolumn{4}{c}{Loihi} \\ \cline{3-6} \cline{8-11} 
& & MLP   & LeNet & AlexNet & VGGNet &  & MLP    & LeNet  & AlexNet & VGGNet \\ \hline \\[-17pt] \hline
Accuracy (\%) & & 92.24 & 98.93 & 98.68   & 99.60  &  & 90.67  & 93.39  & 95.13 & 97.70 \\
Inference Power (mW) & & 878 & 852 & 857 & 825 & & 61.37 & 58.06 & 79.09  & 71.89 \\
Latency (mS) &  & 2.13 & 2.50 & 2.59   & 2.59  &  & 12.64  & 8.97   & 11.40   & 8.03   \\
Inference Energy (mJ) & & 1.87 & 2.13 & 2.22   & 2.14  &  & 0.78  & 0.52  & 0.90   & 0.58  \\ \hline \hline
\end{tabular}
}
\label{table:Comprehensive_Alphabets}
\end{table}

\begin{table}[H]
\caption{Movidious vs Loihi - Comprehensive Analysis - ASL Digits}
\centering
\renewcommand{\arraystretch}{1.2}
\resizebox{\columnwidth}{!}{%
\begin{tabular}{clcccclcccc}
\hline \hline
\multirow{2}{*}{\begin{tabular}[c]{@{}c@{}}Benchmarking\\ Metric\end{tabular}} &  & \multicolumn{4}{c}{NCS2}     &  & \multicolumn{4}{c}{Loihi}          \\ \cline{3-6} \cline{8-11} 
&  & MLP   & LeNet & AlexNet & VGGNet &  & MLP    & LeNet  & AlexNet & VGGNet \\ \hline \\ [-17pt] \hline
Accuracy (\%) &  & 86.19 & 98.55 & 99.03   & 99.03  &  & 85.47  & 95.40  & 97.34   & 98.06  \\
Inference Power (mW) &  & 849   & 838   & 837     & 840    &  & 60.23 & 60.53 & 90.34  & 84.54 \\
Latency (mS) &  & 2.06 & 2.29 & 2.39   & 2.47  &  & 16.29  & 9.67   & 9.04    & 7.77   \\
Inference Energy (mJ) &  & 1.75 & 1.92 & 2.00   & 2.08  &  & 0.98  & 0.59  & 0.82   & 0.66  \\ \hline \hline
\end{tabular}
}
\label{table:Comprehensive_Digits}
\end{table}

\subsubsection{Increasing Data Sparseness}

\textcolor{black}{Our experiments thus far have employed ANN models trained on grayscale images of the ASL datasets. However, in some cases, all of the information in these grayscale images may not be necessary to classify the hand gestures. Thus, using similar edge detection and methods to \cite{chandarana_2021}, we reduce the information in the images by increasing their sparseness. We then re-trained and converted the models using the edge detected images using a fixed duration of 120 ms and report the metrics in Table \ref{tab:sparse} for the ASL Alphabet.}

\begin{table}[H]
\centering
\color{black}
\caption{\textcolor{black}{NCS2 vs Loihi performance using edge detected images (ASL-Alphabet)}}
\label{tab:sparse}
\resizebox{\textwidth}{!}{
\begin{tabular}{ccccccccccc}
\hline
\multirow{2}{*}{\begin{tabular}[c]{@{}c@{}}Benchmarking\\ Metrics\end{tabular}} &  & \multicolumn{4}{c}{NCS2} &  & \multicolumn{4}{c}{Loihi} \\ \cline{3-6} \cline{8-11} 
 &  & MLP & LeNet & AlexNet & VGGNet &  & MLP & LeNet & AlexNet & VGGNet \\ \hline
Accuracy (\%) &  & 87.72 & 87.93 & 90.69 & 95.18 &  & 86.44 & 64.91 & 88.64 & 92.78 \\
Inference Power (mW) &  & 863 & 847 & 868 & 853 &  & 42.21 & 41.03 & 62.93 & 61.55 \\
Latency (mS) &  & 2.05 & 2.35 & 2.51 & 2.59 &  & 14.26 & 12.84 & 19.28 & 16.93 \\
Inference Energy (mJ) &  & 1.76 & 1.99 & 2.18 & 2.21 &  & 0.60 & 0.53 & 1.21 & 1.04 \\ \hline
\end{tabular}}
\end{table}

\textcolor{black}{Comparing Tables \ref{table:Comprehensive_Alphabets} and \ref{tab:sparse}, we see that, in general, using edge detection on the data reduces the power of the SNN by $14.38\%$ to $31.22\%$ compared to the original grayscale images on Loihi. As we can see in Table \ref{tab:sparse}, most of the models have higher energy consumption. This is due to the increase in latency resulting from fixing the duration to a constant 120 ms between models. We can also see that the Intel NCS2 does not benefit from the increased sparseness of the images. This results in an even greater difference of $13.79\times$ to $20.64\times$ in power consumption between the Intel NCS2 and Loihi, respectively.}

\section{Conclusion}

In this work, we first trained four different ANN models on two static ASL hand gesture image datasets, the ASL Alphabet and ASL Digits. We then modified, constrained, and trained versions of four common ANN models to ensure compatibility with the SNN Conversion Toolbox. We then deployed the converted SNNs on Intel's neuromorphic processor, Loihi, and benchmarked them against their conventional ANN implementations on the Intel NCS2 edge AI accelerator herein. We then performed an analysis of the correlation between the accuracy and duration/latency of the SNN models and provided a method to find a balanced point according to an application's tolerance to accuracy or latency. Moreover, we discussed specific mechanisms to accurately measure the power consumption and latency in both neuromorphic and edge computing hardware. Finally, we provided a comprehensive comparison between Loihi and NCS2 in terms of accuracy, latency, power consumption, and energy. In terms of accuracy, the SNN models approach or even surpass ANN models. In terms of latency, Loihi falls behind Intel's NCS2, however, the power reduction realized by Loihi is so significant (9.26-20.64$\times$) that it could still achieve 1.79-4.10$\times$ energy savings compared to NCS2 while executing the ASL classification tasks. Future work includes dynamic hand gesture recognition of ASL using temporal learning rules for SNNs.

\appendix
\section{Duration vs. Latency}
\label{sec:duration_latency_lines}
\begin{figure}[H]
    \centering
    \subfloat[]{
        \includegraphics[width=.45\textwidth]{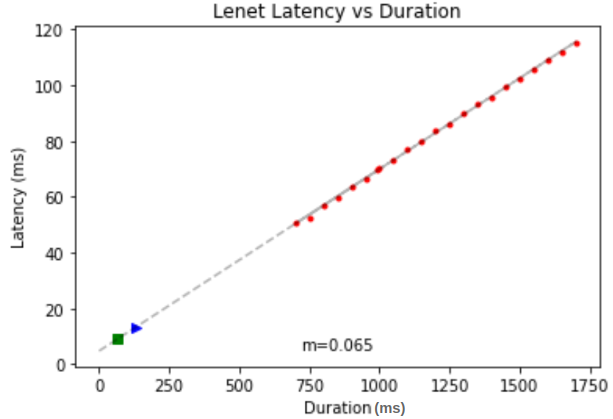}
    }
    \subfloat[]{
        \includegraphics[width=.45\textwidth]{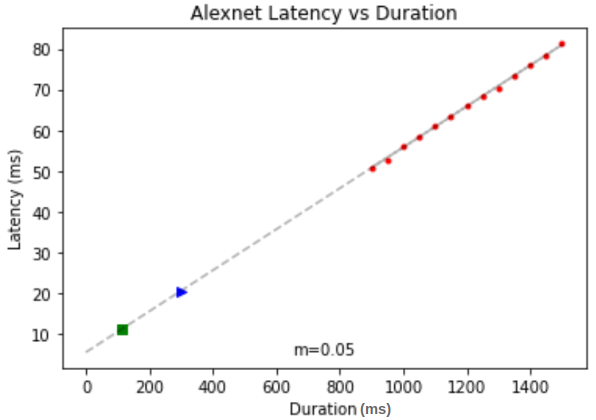}
    }
    \newline
    \subfloat[]{
        \includegraphics[width=.45\textwidth]{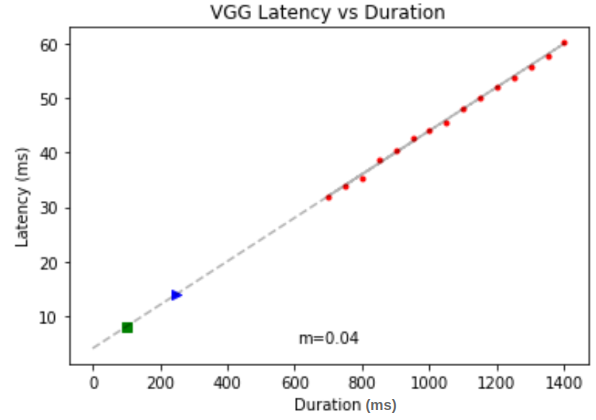}
    }
    \caption{ASL Alphabet Latency vs. Duration. Legend: \shapelabel{circle}{red}{} test points, \shapelabel{regular polygon,regular polygon sides=3}{blue}{} best accuracy, \shapelabel{regular polygon,regular polygon sides=4}{black!30!green}{} balanced duration-accuracy point.}
    \label{fig:latency_duration_alpha}
\end{figure}

\begin{figure}[H]
    \centering
    \subfloat[]{
        \includegraphics[width=.45\textwidth]{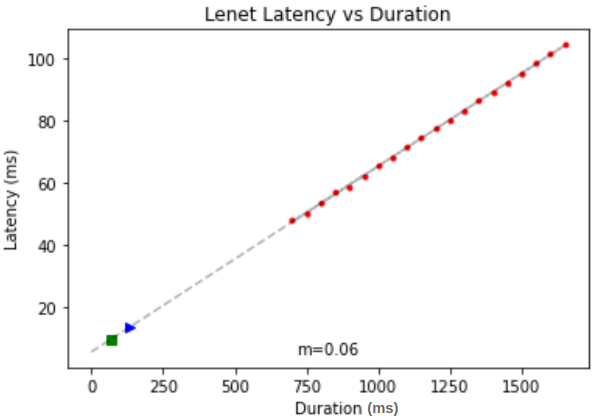}
    }
    \subfloat[]{
        \includegraphics[width=.45\textwidth]{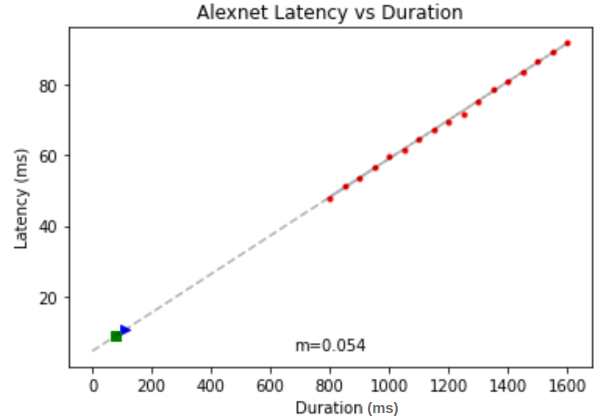}
    }
    \newline
    \subfloat[]{
        \includegraphics[width=.45\textwidth]{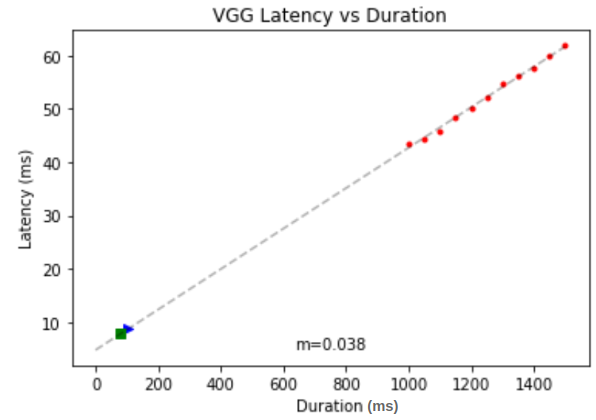}
    }
    \caption{ASL Digits Latency vs. Duration. Legend: \shapelabel{circle}{red}{} test points, \shapelabel{regular polygon,regular polygon sides=3}{blue}{} best accuracy, \shapelabel{regular polygon,regular polygon sides=4}{black!30!green}{} balanced duration-accuracy point.}
    \label{fig:latency_duration_digit}
\end{figure}

\section{Confusion Matrices}
\label{sec:confusion_matrices}

\textcolor{black}{To visualize how the ANN and SNN differ on a class-by-class basis we have included difference confusion matrices for each of the models in our experiments. These figures are dubbed ``difference'' confusion matrices because they represent the difference between two models. When the color value of a pixel is close to zero on the color scale, this conveys that the SNN and ANN or C-ANN were about equal in their performance. The diagonals of these difference matrices are emphasized with white asterisks to show where the SNN and ANN correctly classified the input. If a pixel on the diagonal is bright yellow then the SNN correctly classified more inputs than the ANN or C-ANN. However, if the pixel on the diagonal is dark blue or purple the ANN or C-ANN was more accurate than the SNN at classifying the input. If the pixel does not lie on the diagonal of the matrix and is yellow then this concludes that the SNN was more incorrect than the ANN or C-ANN and vice-versa.}
\begin{figure}[H]
    \centering
        \includegraphics[width=.4\textwidth]{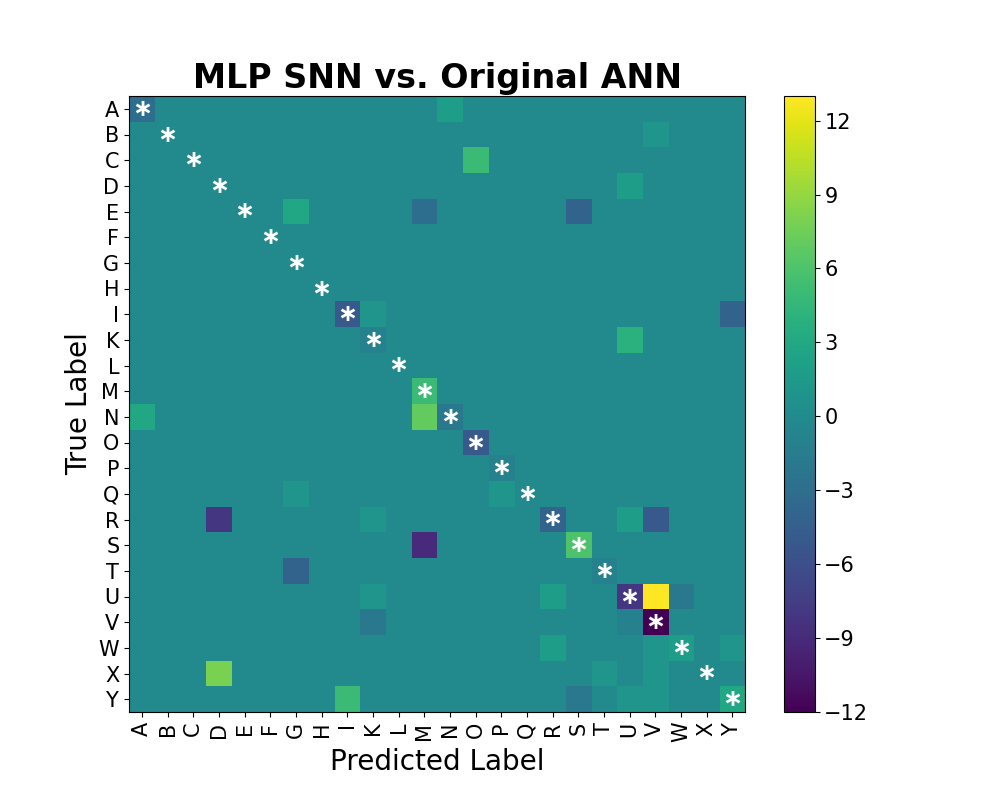}
        \includegraphics[width=.4\textwidth]{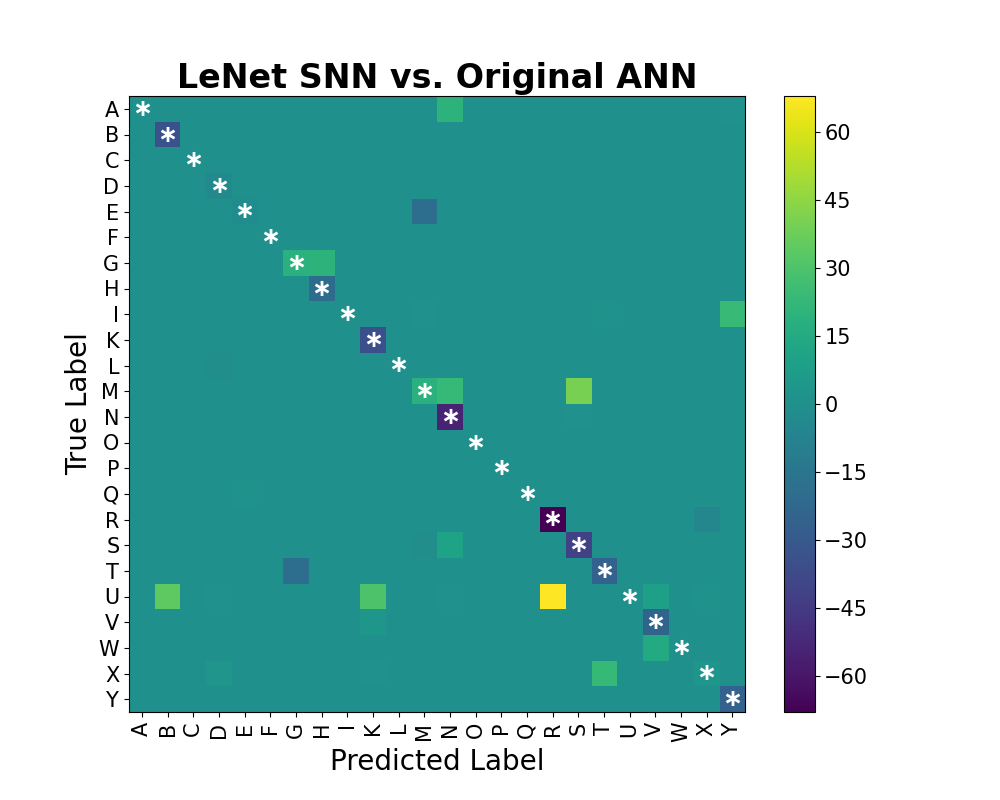}
        \includegraphics[width=.4\textwidth]{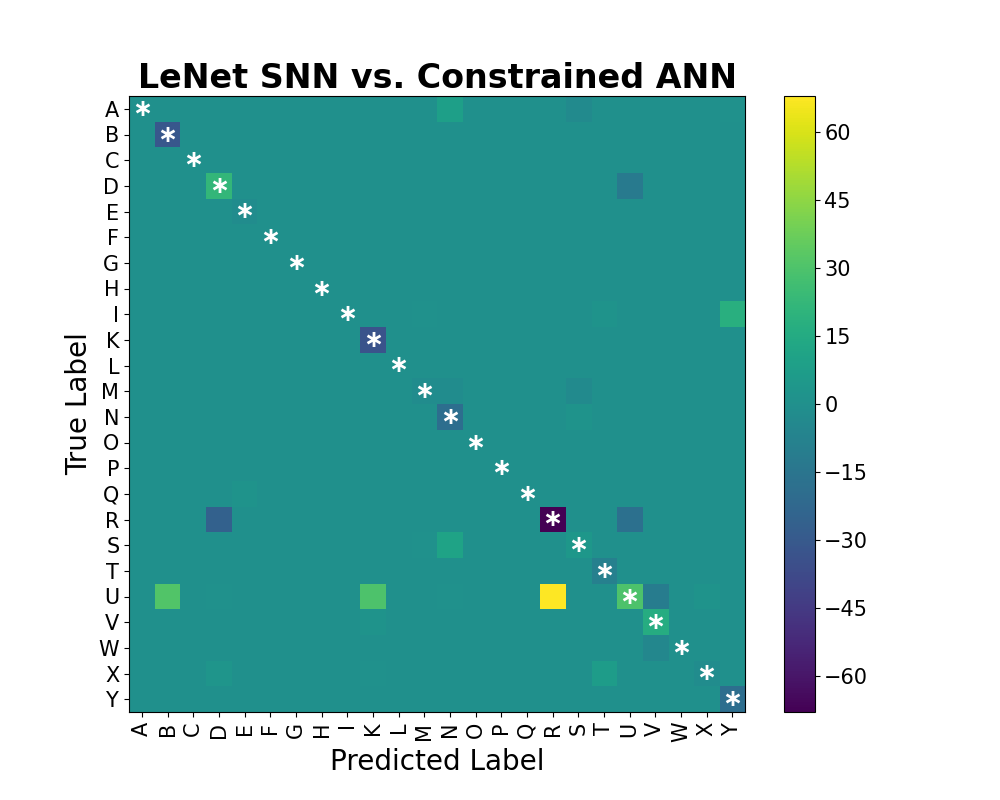}
        \includegraphics[width=.4\textwidth]{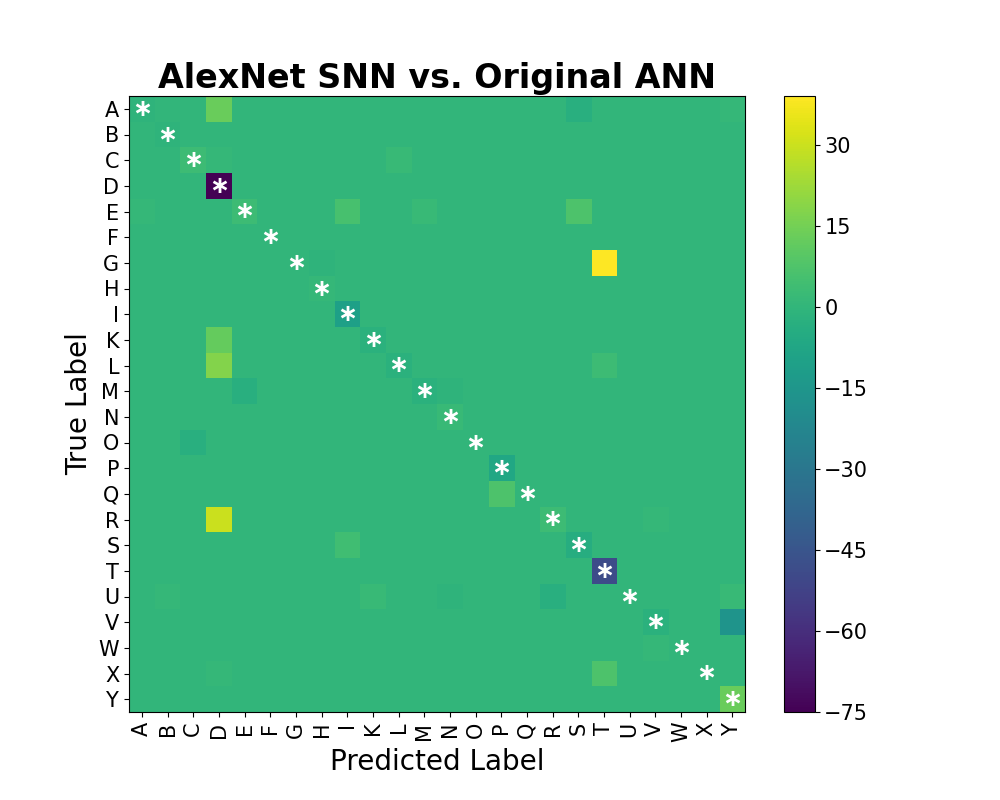}
        \includegraphics[width=.4\textwidth]{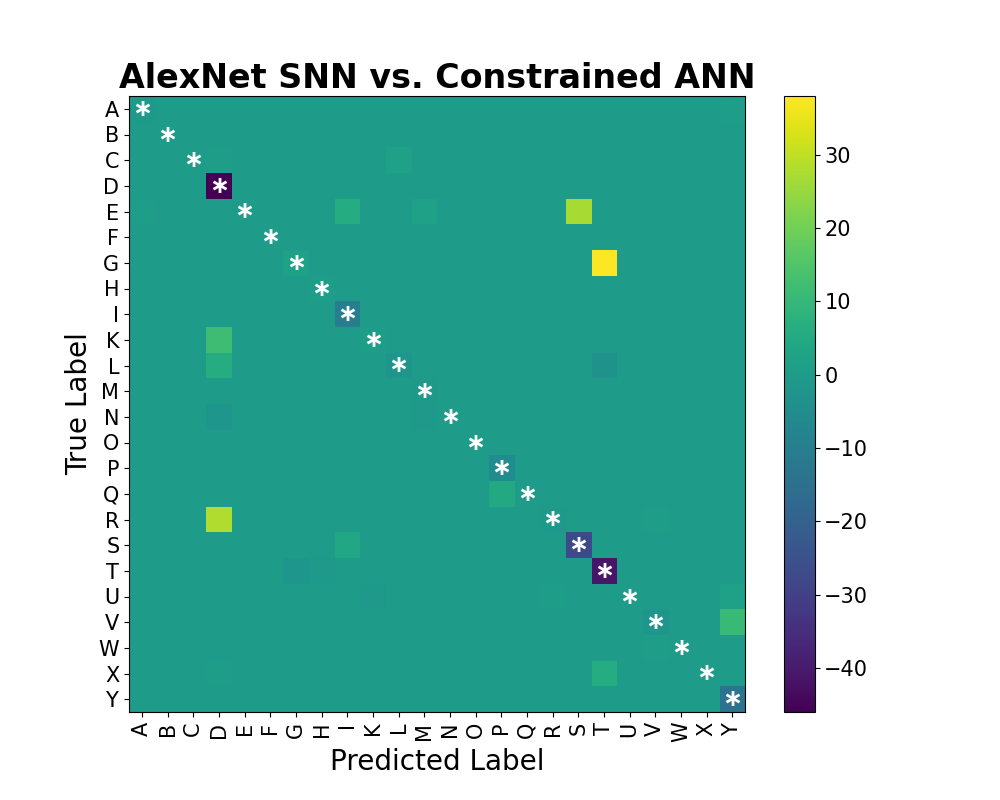}
        \includegraphics[width=.4\textwidth]{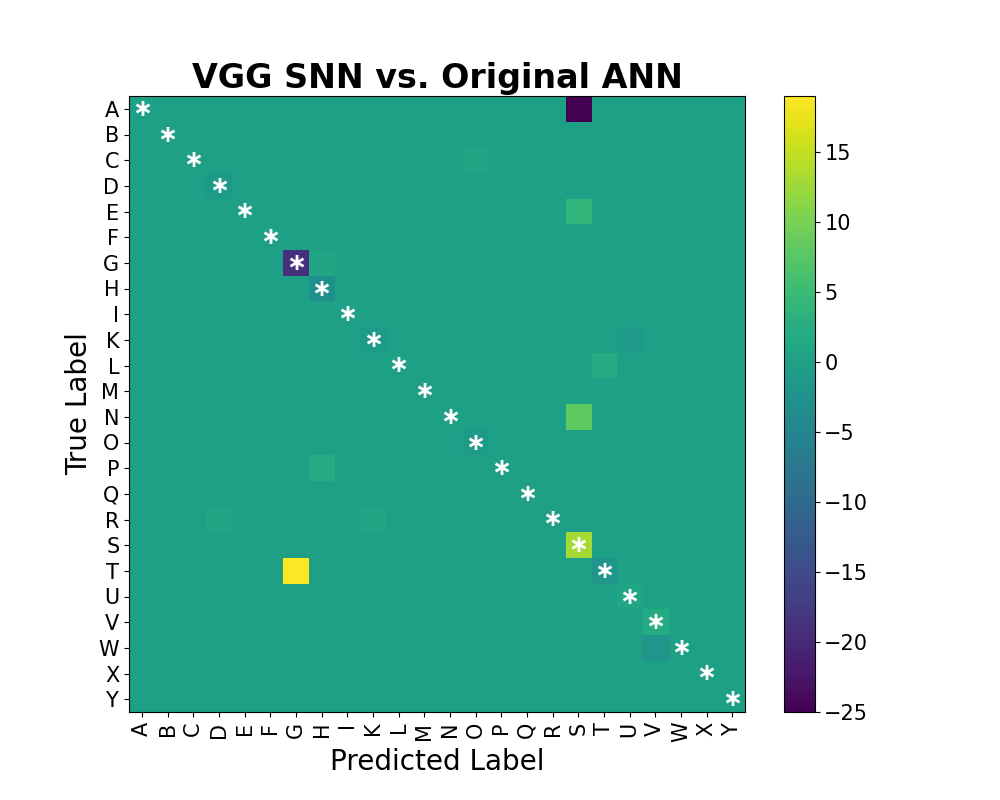}
        \includegraphics[width=.4\textwidth]{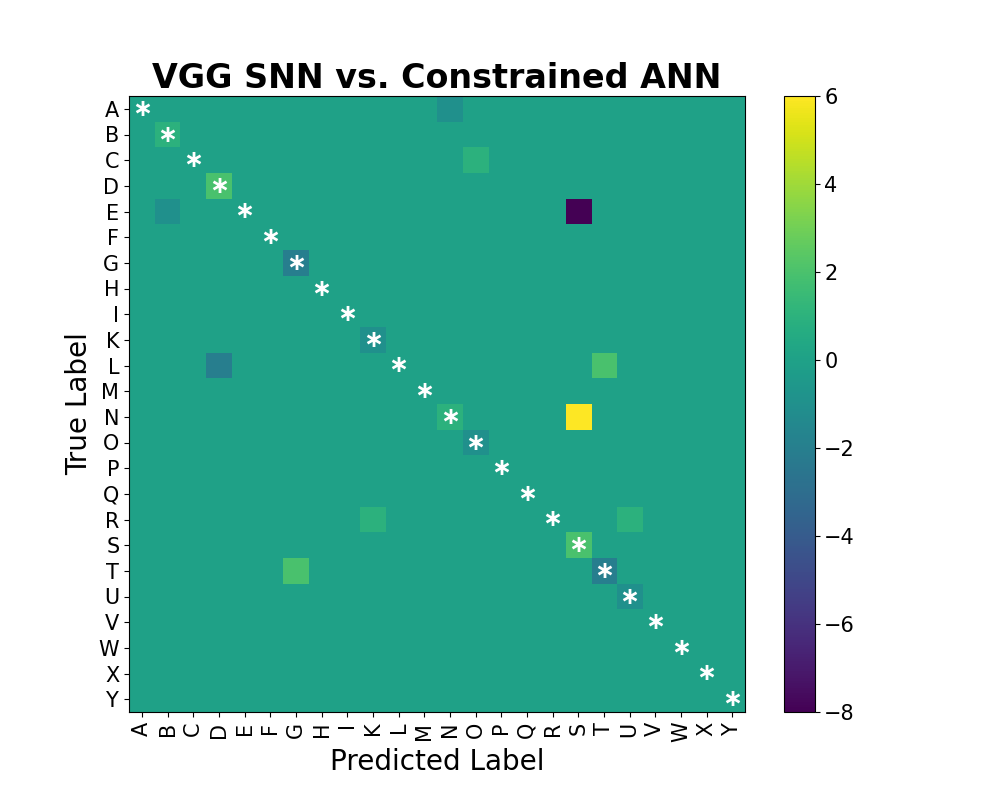}
    \caption{ASL Alphabet Difference Confusion Matrices.}
    \label{fig:alpha_confusion_matrices}
\end{figure}

\begin{figure}[H]
    \centering
        \includegraphics[width=.4\textwidth]{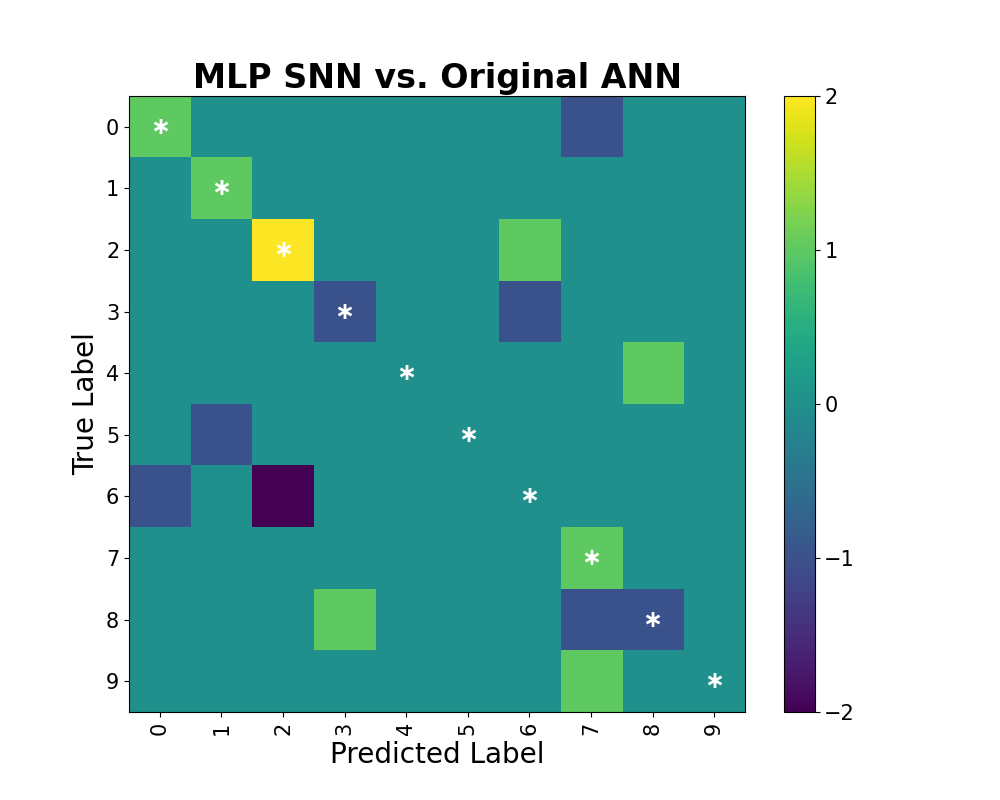}
        \includegraphics[width=.4\textwidth]{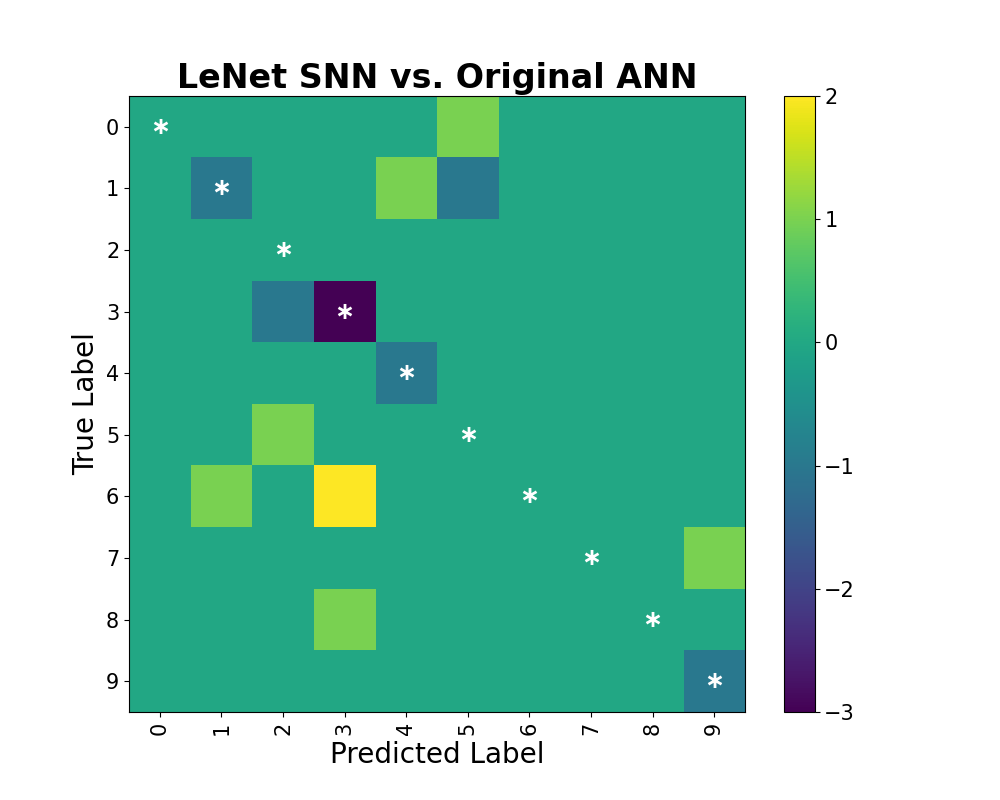}
        \includegraphics[width=.4\textwidth]{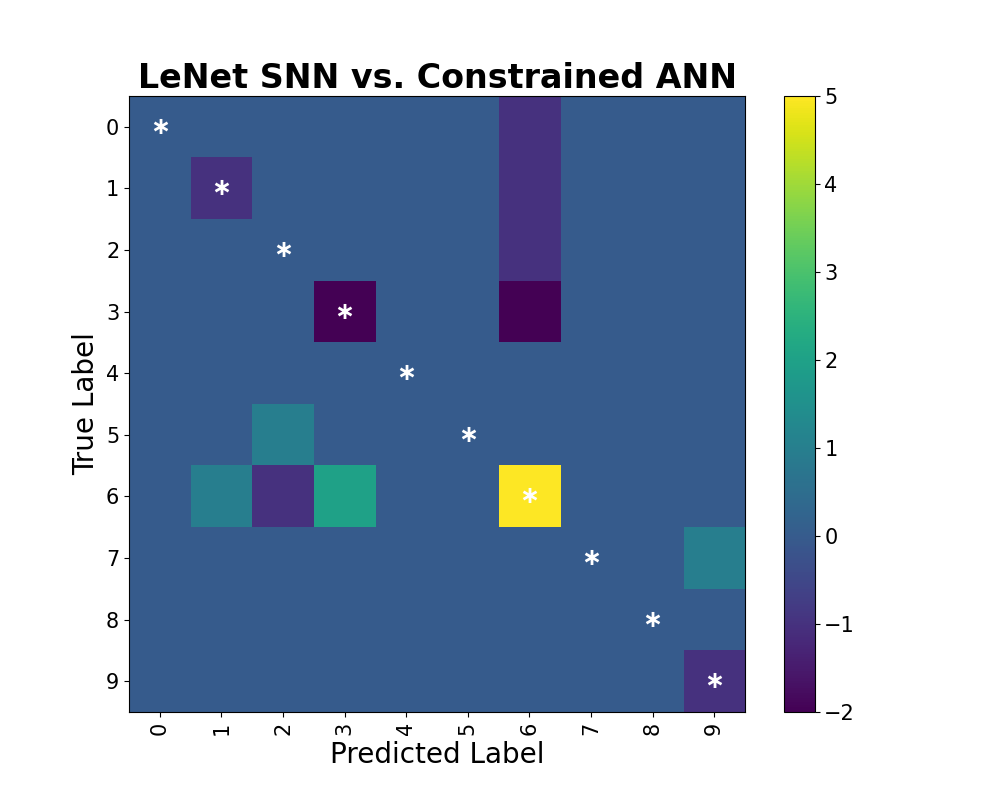}
        \includegraphics[width=.4\textwidth]{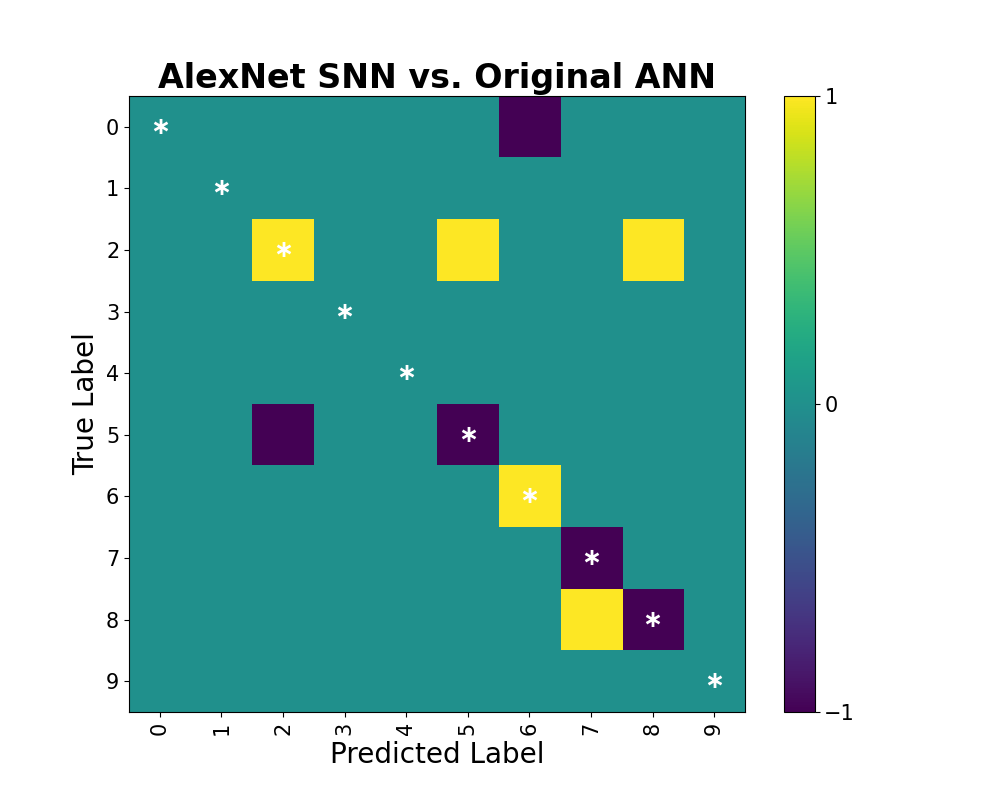}
        \includegraphics[width=.4\textwidth]{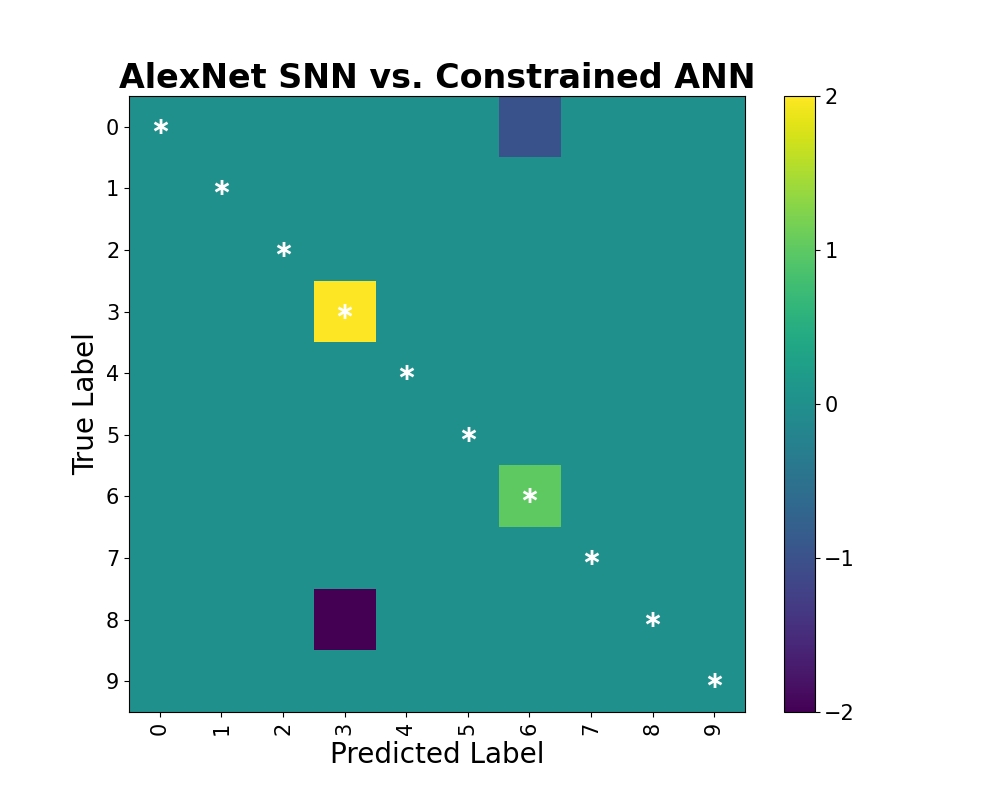}
        \includegraphics[width=.4\textwidth]{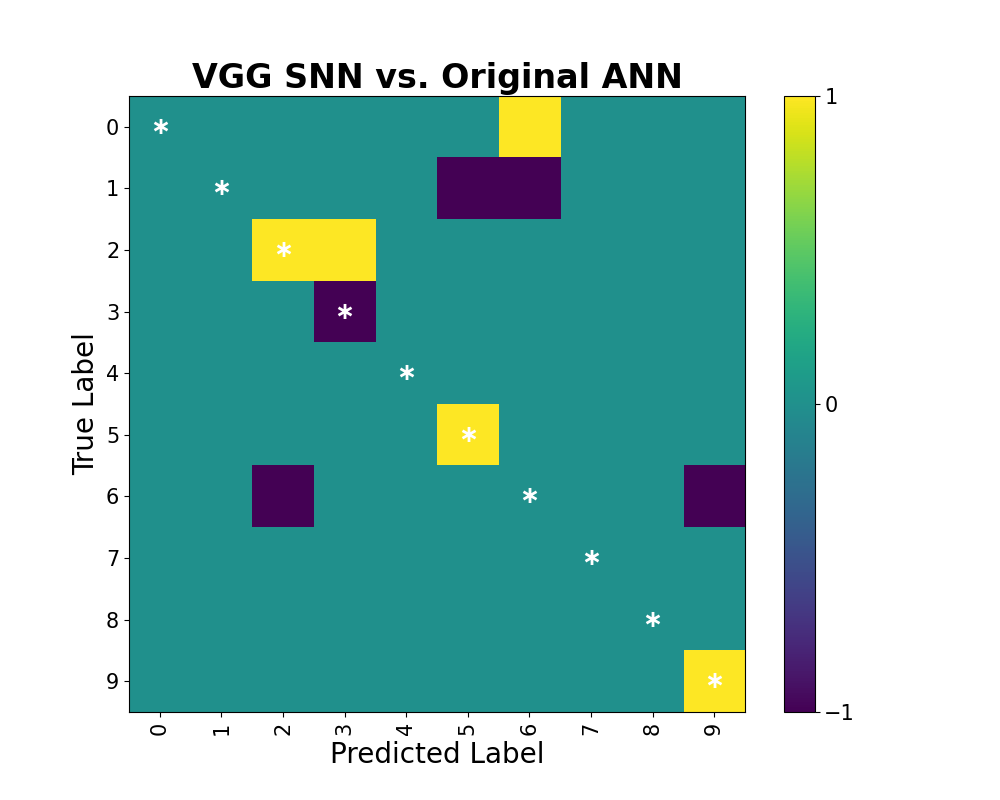}
        \includegraphics[width=.4\linewidth]{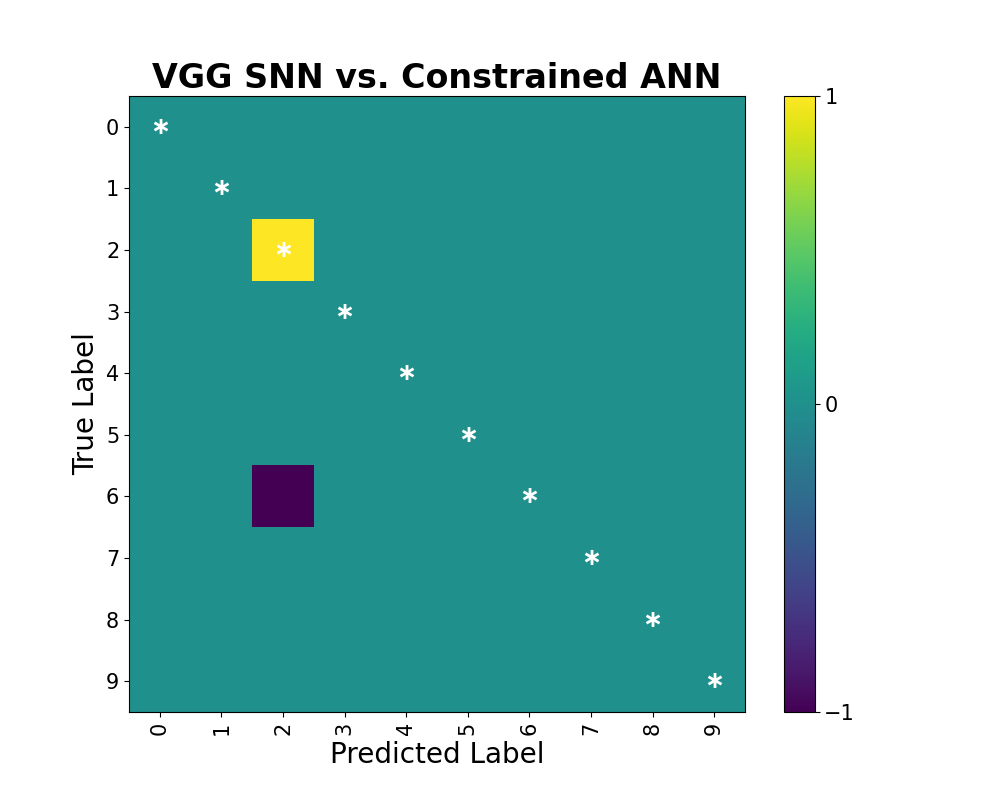}
    \caption{ASL Digits Difference Confusion Matrices.}
    \label{fig:digit_confusion_matrices}
\end{figure}

\section{Parameter Size Effect on Latency and Power}
\label{appendix:parameters_lenet}
\textcolor{black}{To examine the impact of varying parameter sizes on Loihi's latency and power consumption, we trained three LeNet models. Each of these models have the standard depth of 5 layers but differ in the number of kernels each of the convolution layers use.
According to Table \ref{table:lenet_varying_params}, increasing the number of parameters increases power consumption and latency. Consequently, the number of network parameters has a direct impact on both latency and power consumption.}

\begin{table}[H]
\centering
\color{black}
\caption{\textcolor{black}{ASL Alphabet LeNet - Varying parameter size comparison on Loihi.}}
\label{table:lenet_varying_params}
\begin{tabular}{ccccc}
\hline
\textbf{Model Size} & \textbf{Parameters} & \textbf{Power (mW)} & \textbf{Latency (ms)} & \textbf{Energy (mJ)} \\ \hline
Small & 22870 & 35.64 & 12.11 & 0.432 \\
Medium & 45616 & 43.33 & 12.36 & 0.535 \\
Large & 91258 & 54.64 & 19.24 & 1.051 \\ \hline
\end{tabular}
\end{table}

\section{Accuracy Confidence Intervals}
\textcolor{black}{For all models, we calculated the confidence intervals \cite{nowotny2014two} with a confidence level of 95\%. As shown, the uncertainty margin is smaller for the VGGNet ANN compared to the others. However, as we can see, every model has a narrow margin of uncertainty.}

\begin{figure}[H]
         \centering
         \includegraphics[width=\textwidth]{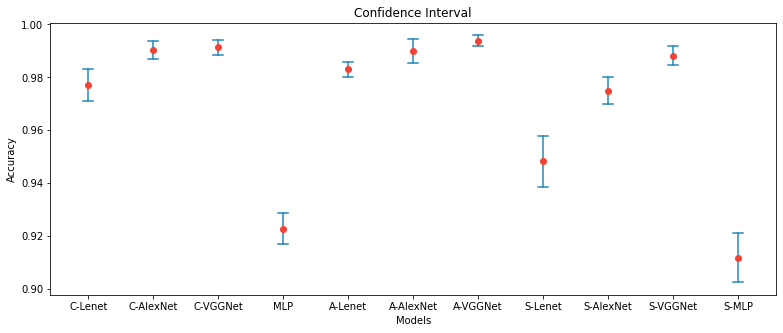}
    \caption{\textcolor{black}{Confidence interval plot for different models on ASL Alphabet dataset.}}
     \label{fig:CI_A}
\end{figure}

\begin{figure}[H]
         \centering
         \includegraphics[width=\textwidth]{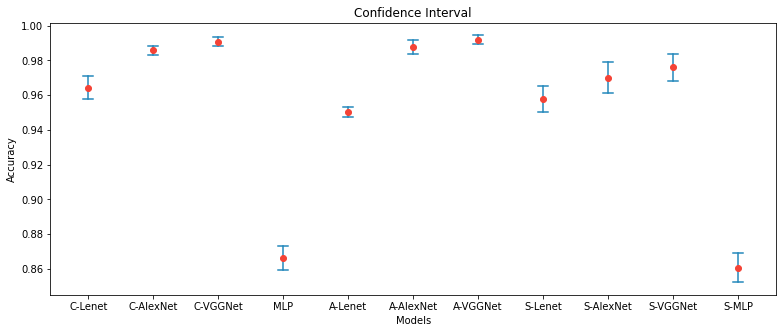}
    \caption{\textcolor{black}{Confidence interval plot for different models on ASL Digit dataset.}}
     \label{fig:CI_D}
\end{figure}
\textcolor{black}{
To compute the confidence intervals in Figures \ref{fig:CI_A} and \ref{fig:CI_D}, we use the following equation \cite{petty2012calculating}:
\begin{equation}
    \centering
    CI = Mean \pm Z_{score} * \frac{Sd}{\sqrt{N}}
\end{equation}
where $Sd$ is the standard deviation and $N$ is the number of samples. In Figures \ref{fig:CI_A} and \ref{fig:CI_D}, we illustrate confidence intervals of 95\% which correlate to a z-score of 1.96. Herein, we assume a Gaussian prior in our confidence interval calculations.}

\section*{Acknowledgment}
This work is partially supported by an ASPIRE grant from the Office of the Vice President for Research at the University of South Carolina. Special thanks to the Intel Neuromorphic Research Community (INRC) for providing access to the Loihi chips for the experiments performed in this paper. 

\section*{References}
\bibliographystyle{iopart-num}
\bibliography{refs}

\providecommand{\newblock}{}
\begin{thebibliography}{10}
\expandafter\ifx\csname url\endcsname\relax
  \def\url#1{{\tt #1}}\fi
\expandafter\ifx\csname urlprefix\endcsname\relax\def\urlprefix{URL }\fi
\providecommand{\eprint}[2][]{\url{#2}}

\bibitem{cheok2019review}
Cheok M~J, Omar Z and Jaward M~H 2019 {\em International Journal of Machine
  Learning and Cybernetics\/} {\bf 10} 131--153

\bibitem{tolentino2019static}
Tolentino L~K~S, Juan R~O~S, Thio-ac A~C, Pamahoy M~A~B, Forteza J~R~R and
  Garcia X~J~O 2019 {\em International Journal of Machine Learning and
  Computing\/} {\bf 9} 821--827

\bibitem{liao2019dynamic}
Liao Y, Xiong P, Min W, Min W and Lu J 2019 {\em IEEE Access\/} {\bf 7}
  38044--38054

\bibitem{wadhawan2021sign}
Wadhawan A and Kumar P 2021 {\em Archives of Computational Methods in
  Engineering\/} {\bf 28} 785--813

\bibitem{shukor2015new}
Shukor A~Z, Miskon M~F, Jamaluddin M~H, bin Ali F, Asyraf M~F, bin Bahar M~B
  {\em et~al.\/} 2015 {\em Procedia Computer Science\/} {\bf 76} 60--67

\bibitem{kumar2017real}
Kumar P, Saini R, Behera S~K, Dogra D~P and Roy P~P 2017 {\em 2017 Fifteenth
  IAPR international conference on machine vision applications (MVA)\/} (IEEE)
  pp 157--160

\bibitem{zhang2011framework}
Zhang X, Chen X, Li Y, Lantz V, Wang K and Yang J 2011 {\em IEEE Transactions
  on Systems, Man, and Cybernetics-Part A: Systems and Humans\/} {\bf 41}
  1064--1076

\bibitem{hernandez2004new}
Hernandez-Rebollar J~L, Kyriakopoulos N and Lindeman R~W 2004 {\em Sixth IEEE
  International Conference on Automatic Face and Gesture Recognition, 2004.
  Proceedings.\/} (IEEE) pp 547--552

\bibitem{bui2007recognizing}
Bui T~D and Nguyen L~T 2007 {\em IEEE sensors journal\/} {\bf 7} 707--712

\bibitem{elbadawy2015proposed}
ElBadawy M, Elons A~S, Sheded H and Tolba M~F 2015 {\em Intelligent Systems'
  2014\/} (Springer) pp 721--730

\bibitem{7025313}
Marin G, Dominio F and Zanuttigh P 2014 {\em 2014 IEEE International Conference
  on Image Processing (ICIP)\/} pp 1565--1569

\bibitem{kumar2017coupled}
Kumar P, Gauba H, Roy P~P and Dogra D~P 2017 {\em Pattern Recognition
  Letters\/} {\bf 86} 1--8

\bibitem{garcia2016real}
Garcia B and Viesca S~A 2016 {\em Convolutional Neural Networks for Visual
  Recognition\/} {\bf 2} 225--232

\bibitem{rautaray2015vision}
Rautaray S~S and Agrawal A 2015 {\em Artificial intelligence review\/} {\bf 43}
  1--54

\bibitem{pisharady2015recent}
Pisharady P~K and Saerbeck M 2015 {\em Computer Vision and Image
  Understanding\/} {\bf 141} 152--165

\bibitem{d2016recent}
D’Orazio T, Marani R, Ren{\`o} V and Cicirelli G 2016 {\em Image and Vision
  Computing\/} {\bf 52} 56--72

\bibitem{rastgoo2018multi}
Rastgoo R, Kiani K and Escalera S 2018 {\em Entropy\/} {\bf 20} 809

\bibitem{adithya2020deep}
Adithya V and Rajesh R 2020 {\em Procedia Computer Science\/} {\bf 171}
  2353--2361

\bibitem{barbhuiya2021cnn}
Barbhuiya A~A, Karsh R~K and Jain R 2021 {\em Multimedia Tools and
  Applications\/} {\bf 80} 3051--3069

\bibitem{rahman2019new}
Rahman M~M, Islam M~S, Rahman M~H, Sassi R, Rivolta M~W and Aktaruzzaman M 2019
  {\em 2019 International Conference on Sustainable Technologies for Industry
  4.0 (STI)\/} (IEEE) pp 1--6

\bibitem{verhelst2017embedded}
Verhelst M and Moons B 2017 {\em IEEE Solid-State Circuits Magazine\/} {\bf 9}
  55--65

\bibitem{schuman2022opportunities}
Schuman C~D, Kulkarni S~R, Parsa M, Mitchell J~P, Kay B {\em et~al.\/} 2022
  {\em Nature Computational Science\/} {\bf 2} 10--19

\bibitem{loihi_2021}
Davies M, Wild A, Orchard G, Sandamirskaya Y, Guerra G~A~F, Joshi P, Plank P
  and Risbud S~R 2021 {\em Proceedings of the IEEE\/} {\bf 109} 911--934

\bibitem{loihi_2018}
Davies M, Srinivasa N, Lin T~H, Chinya G, Cao Y, Choday S~H, Dimou G, Joshi P,
  Imam N, Jain S, Liao Y, Lin C~K, Lines A, Liu R, Mathaikutty D, McCoy S, Paul
  A, Tse J, Venkataramanan G, Weng Y~H, Wild A, Yang Y and Wang H 2018 {\em
  IEEE Micro\/} {\bf 38} 82--99

\bibitem{snntoolbox}
Rueckauer B, Lungu I~A, Hu Y, Pfeiffer M and Liu S~C 2017 {\em Frontiers in
  Neuroscience\/} {\bf 11} ISSN 1662-453X
  \urlprefix\url{https://www.frontiersin.org/article/10.3389/fnins.2017.00682}

\bibitem{perez_conversion}
Pérez-Carrasco J~A, Zhao B, Serrano C, Acha B, Serrano-Gotarredona T, Chen S
  and Linares-Barranco B 2013 {\em IEEE Transactions on Pattern Analysis and
  Machine Intelligence\/} {\bf 35} 2706--2719

\bibitem{cao_conversion}
Cao Y, Chen Y and Khosla D 2015 {\em International Journal of Computer
  Vision\/} {\bf 113} 54--66
  \urlprefix\url{https://doi.org/10.1007/s11263-014-0788-3}

\bibitem{tensorflow}
Abadi M, Barham P, Chen J, Chen Z, Davis A {\em et~al.\/} 2016 {\em 12th
  {USENIX} Symposium on Operating Systems Design and Implementation ({OSDI}
  16)\/} (Savannah, GA: {USENIX} Association) pp 265--283 ISBN
  978-1-931971-33-1

\bibitem{pytorch}
Paszke A, Gross S, Massa F, Lerer A, Bradbury J, Chanan G, Killeen T, Lin Z,
  Gimelshein N, Antiga L, Desmaison A, Kopf A, Yang E, DeVito Z, Raison M,
  Tejani A, Chilamkurthy S, Steiner B, Fang L, Bai J and Chintala S 2019 {\em
  Advances in Neural Information Processing Systems 32\/} ed Wallach H,
  Larochelle H, Beygelzimer A, d\textquotesingle Alch\'{e}-Buc F, Fox E and
  Garnett R (Curran Associates, Inc.) pp 8024--8035

\bibitem{nxtf}
Rueckauer B, Bybee C, Goettsche R, Singh Y, Mishra J and Wild A 2021 Nxtf: An
  api and compiler for deep spiking neural networks on intel loihi
  (\textit{Preprint} \eprint{2101.04261})

\bibitem{signlangmnistkaggle}
Sign language mnist
  \urlprefix\url{https://www.kaggle.com/datamunge/sign-language-mnist}

\bibitem{mavi2020new}
Mavi A 2020 {\em arXiv preprint arXiv:2011.08927\/}

\bibitem{lecun2010mnist}
LeCun Y, Cortes C and Burges C 2010 {\em ATT Labs [Online]. Available:
  http://yann.lecun.com/exdb/mnist\/} {\bf 2}

\bibitem{lecun1998gradient}
LeCun Y, Bottou L, Bengio Y and Haffner P 1998 {\em Proceedings of the IEEE\/}
  {\bf 86} 2278--2324

\bibitem{krizhevsky2012imagenet}
Krizhevsky A, Sutskever I and Hinton G~E 2012 {\em Advances in neural
  information processing systems\/} {\bf 25} 1097--1105

\bibitem{simonyan2014very}
Simonyan K and Zisserman A 2014 {\em arXiv preprint arXiv:1409.1556\/}

\bibitem{tensorflow2015-whitepaper}
Abadi M, Agarwal A, Barham P, Brevdo E, Chen Z, Citro C, Corrado G~S, Davis A,
  Dean J, Devin M, Ghemawat S, Goodfellow I, Harp A, Irving G, Isard M, Jia Y,
  Jozefowicz R, Kaiser L, Kudlur M, Levenberg J, Man\'{e} D, Monga R, Moore S,
  Murray D, Olah C, Schuster M, Shlens J, Steiner B, Sutskever I, Talwar K,
  Tucker P, Vanhoucke V, Vasudevan V, Vi\'{e}gas F, Vinyals O, Warden P,
  Wattenberg M, Wicke M, Yu Y and Zheng X 2015 {TensorFlow}: Large-scale
  machine learning on heterogeneous systems software available from
  tensorflow.org \urlprefix\url{https://www.tensorflow.org/}

\bibitem{massa2020efficient}
Massa R, Marchisio A, Martina M and Shafique M 2020 {\em 2020 International
  Joint Conference on Neural Networks (IJCNN)\/} (IEEE) pp 1--9

\bibitem{buettner2021heartbeat}
Buettner K and George A~D 2021 {\em 2021 IEEE Computer Society Annual Symposium
  on VLSI (ISVLSI)\/} (IEEE) pp 138--143

\bibitem{Intel:2019}
Intel 2019 Intel neural compute stick 2
  \urlprefix\url{https://www.intel.com/content/www/us/en/developer/
  tools/neural-compute-stick/overview.html}

\bibitem{WinNT}
Amazon.2019.makerhawk um34
  \urlprefix\url{https://www.amazon.com/MakerHawkBluetooth-Voltmeter-
  Multimeter-Resistance/dp/B07DK4GDSP}

\bibitem{loihi_2018_mapping}
Lin C~K, Wild A, Chinya G~N, Lin T~H, Davies M and Wang H 2018 {\em SIGPLAN
  Not.\/} {\bf 53} 78–89 ISSN 0362-1340
  \urlprefix\url{https://doi.org/10.1145/3296979.3192371}

\bibitem{cuba}
Vogels T~P and Abbott L~F 2005 {\em Journal of Neuroscience\/} {\bf 25}
  10786--10795 ISSN 0270-6474 (\textit{Preprint}
  \eprint{https://www.jneurosci.org/content/25/46/10786.full.pdf})
  \urlprefix\url{https://www.jneurosci.org/content/25/46/10786}

\bibitem{neuronal_dynamics}
Gerstner W, Kistler W~M, Naud R and Paninski L 2014 {\em Neuronal Dynamics:
  From single neurons to networks and models of cognition\/} (Cambridge
  University Press)

\bibitem{blouw2019benchmarking}
Blouw P, Choo X, Hunsberger E and Eliasmith C 2019 {\em Proceedings of the 7th
  Annual Neuro-inspired Computational Elements Workshop\/} pp 1--8

\bibitem{mannan2022hypertuned}
Mannan A, Abbasi A, Javed A~R, Ahsan A, Gadekallu T~R and Xin Q 2022 {\em
  Computational Intelligence and Neuroscience\/} {\bf 2022}

\bibitem{can2021deep}
Can C, Kaya Y and K{\i}l{\i}{\c{c}} F 2021 {\em Biomedical Physics \&
  Engineering Express\/} {\bf 7} 055005

\bibitem{fregoso2021optimization}
Fregoso J, Gonzalez C~I and Martinez G~E 2021 {\em Axioms\/} {\bf 10} 139

\bibitem{xiao2022sign}
Xiao H, Yang Y, Yu K, Tian J, Cai X, Muhammad U and Chen J 2022 {\em Journal of
  Ambient Intelligence and Humanized Computing\/} {\bf 13} 2131--2141

\bibitem{ncl_models}
\urlprefix\url{https://github.com/intel-nrc-ecosystem/models/tree/master/nxsdk\_modules\_ncl}

\bibitem{snn_coding_2021}
Guo W, Fouda M~E, Eltawil A~M and Salama K~N 2021 {\em Frontiers
  Neuroscience\/}

\bibitem{chandarana_2021}
Chandarana P, Ou J and Zand R 2021 {\em 2021 12th International Green and
  Sustainable Computing Conference (IGSC)\/} pp 1--8

\bibitem{nowotny2014two}
Nowotny T 2014 {\em Frontiers in Robotics and AI\/} {\bf 1} 5

\bibitem{petty2012calculating}
Petty M~D 2012 {\em Proceedings of the Fall 2012 Simulation Interoperability
  Workshop\/} pp 10--14

\end{thebibliography}

\end{document}